\documentclass[square,numbers,sort&compress]{article}

\usepackage[final]{neurips_2024}
\usepackage{natbib}

\usepackage[utf8]{inputenc} 
\usepackage[T1]{fontenc}    
\usepackage{hyperref}       
\usepackage{url}            
\usepackage{booktabs}       
\usepackage{amsfonts}       
\usepackage{nicefrac}       
\usepackage{microtype}      
\usepackage{xcolor}         
\usepackage{algorithm}
\usepackage{algpseudocode}
\usepackage{wrapfig}
\usepackage{hyperref}

\usepackage{microtype}
\usepackage{graphicx}
\usepackage{subfigure}
\usepackage{amsmath}
\usepackage{amssymb}
\usepackage{mathtools}
\usepackage{amsthm}

\usepackage[capitalize,noabbrev]{cleveref}

\usepackage{multirow}
\usepackage{pifont}
\newcommand{\ours}{\textsc{LoT}\,}
\newcommand{\bx}{\mathbf{x}}
\newcommand{\btheta}{\boldsymbol\theta}
\newcommand{\bphi}{\boldsymbol\phi}

\title{Learning from Teaching Regularization: \\ Generalizable Correlations Should be Easy to Imitate}

\author{%
  Can Jin$^{1*}$
  \And Tong Che$^{2*}$
  \And Hongwu Peng$^{3\dagger}$
  \AND Yiyuan Li$^{4\dagger}$
  \And Dimitris N. Metaxas$^{1\ddagger}$
  \And Marco Pavone$^{5\ddagger}$ \\
  \AND \textnormal{$^{1}$Rutgers University\quad $^{2}$Nvidia Research\quad $^{3}$University of Connecticut} \\
  \textnormal{$^{4}$University of North Carolina at Chapel Hill\quad $^{5}$Stanford University} \\
  \texttt{can.jin@rutgers.edu, tongc@nvidia.com}
}

\begin{document}

\maketitle

\begin{abstract}\label{section_abstract}

Generalization remains a central challenge in machine learning. In this work, we propose \textit{Learning from Teaching} (\textbf{\ours}), a novel regularization technique for deep neural networks to enhance generalization. Inspired by the human ability to capture concise and abstract patterns, we hypothesize that generalizable correlations are expected to be easier to imitate. \ours operationalizes this concept to improve the generalization of the main model with auxiliary student learners. The student learners are trained by the main model and, in turn, provide feedback to help the main model capture more generalizable and imitable correlations. Our experimental results across several domains, including Computer Vision, Natural Language Processing, and methodologies like Reinforcement Learning, demonstrate that the introduction of \ours brings significant benefits compared to training models on the original dataset. The results suggest the effectiveness and efficiency of \ours in identifying generalizable information at the right scales while discarding spurious data correlations, thus making \ours a valuable addition to current machine learning. Code is available at \url{https://github.com/jincan333/LoT}.
\end{abstract}

\footnotetext{$^{*}$Equal contribution, $^{^{\dagger}}$Equal contribution, $^{\ddagger}$Equal advising, Correspondence to: Can Jin \href{mailto:can.jin@rutgers.edu}{<can.jin@rutgers.edu>}, Tong Che \href{mailto:tongc@nvidia.com}{<tongc@nvidia.com>}.}

\section{Introduction}\label{section_introduction}

Improving the generalization performance of models on unseen data is a major challenge in machine learning~\citep{rumelhart1985learning, baum1988size, bartlett1996valid, zhang2017understanding, neyshabur2017exploring}. Despite its significant advances, identifying the most generalizable model within the vast space of potential models remains challenging. Existing deep learning approaches focus on crafting the hypothesis spaces where prediction errors are optimized using training data~\citep{michalski1983theory, haussler1988quantifying, nakkiran2020deep}. These spaces are shaped by inductive biases~\citep{mitchell1980need, haussler1988quantifying} embedded in the neural architectures which include implicit assumptions about the data~\citep{abnar2020transferring, d2021convit, touvron2021training}, objective functions (notably regularizers)~\citep{micchelli2005learning, cortes2012l2, yang2022mixskd}, and learning methodologies~\citep{neyshabur2015path, soudry2018implicit, chaudhari2019entropy}. 

In this paper, to enhance generalization, we use the methodology of regularization~\citep{hinton2012improving, srivastava2014dropout, kukavcka2017regularization}, which prioritizes specific regions in the hypothesis spaces. Regularization techniques often involve employing auxiliary losses or regularizers~\citep{hochreiter1994simplifying, cortes2012l2, yang2022mixskd} alongside the primary task losses. For instance, L1 regularization~\citep{hoerl1970ridge, tibshirani1996regression, tibshirani1997lasso} encourages sparsity within models~\citep{li2017pruning, lee2018snip, hoefler2021sparsity, cho2021espn}. Other regularization techniques include model averaging~\citep{izmailov2018averaging, wortsman2022model}, dropout techniques~\citep{hinton2012improving, wan2013regularization, mianjy2020convergence}, and additional optimization components~\citep{ioffe2015batch, yu2008deep, loshchilov2018decoupled}. Due to its effectiveness and simplicity, regularization is critical in modern machine learning techniques for achieving better generalization~\citep{hinton2012improving, Zheng_2022_WACV}.

We aim to answer the research question: \textit{Among all possible models fitting the training data, which ones are inherently generalizable?} A common belief in cognitive science is that human intelligence development involves distilling information and filtering out extraneous details to discern `simple' correlations among a few selected relevant abstract variables~\citep{tomasello1999human, chopra2019first}. This approach leads to the formation of correlations through simple patterns~\citep{arpit2017closer,li2019ease} at the right scales. However, identifying simple correlations in deep learning remains challenging, mostly due to not being easy to identify the right scale of the problem. Studies in emergent languages suggest that the more structured a language is, the more efficiently it can be transmitted to message receivers~\citep{chaabouni2020compositionality, li2019ease}. Inspired by this finding, we propose defining simple and generalizable correlations at the right scales, as those that can be readily imitated by other learners, provided they possess suitable inductive biases.

Based on this definition, we propose a novel regularization approach, \textit{Learning from Teaching} (\textbf{\ours}). The core of \ours is to compute a measure of `imitability' for the main model to learn data correlations at the correct scales. By adding this measure to the objective function and optimizing it during training, we encourage the teacher model to refine its learned multiscale correlations, making them more accessible through teaching, which in turn leads to better generalization. \ours computes this measure by jointly training the main model as the `teacher' with one or more auxiliary `student' models. The student models strive to distill and assimilate the correlations acquired by the teacher model. Thus, the learning performance of the student defines the measure of imitability of the teacher, which is then used as the \ours regularizer. 

We conduct comprehensive experiments using \ours to improve the Reinforcement Learning (RL) formulation, as well as in Natural Language Processing (NLP) and Computer Vision (CV) applications. In RL, the experimental results demonstrate that \ours attains an average normalized reward enhancement of $44\%$ on four Atari games. In language modeling tasks, \ours achieves significant perplexity reductions on the Penn Tree Bank~\citep{marcus1993building} and WikiText-103~\citep{merity2017pointer}. Notably, \ours enhances the supervised fine-tuning performance of LLaMA~\citep{touvron2023llama, touvron2023llama2} models on GSM8K~\citep{cobbe2021training} and MATH~\citep{hendrycks2021measuring}. In image classification tasks, \ours achieves accuracy gains of $1.99\%$ and $0.83\%$ on CIFAR-100~\citep{krizhevsky2009learning} and ImageNet-1K~\citep{deng2009imagenet}, respectively.

\section{Methodology}
\subsection{Generalizable and Spurious Correlations}\label{Section_3_1}

Given a dataset $\mathcal{D}=\{(\bx_1,y_1), \cdots, (\bx_n,y_n)\}$ generated from a data-generating distribution $\hat{D}$, there are infinitely many continuous functions $f$ such that $f(\bx)=y$ for all $(\bx,y)\in \mathcal{D}$. Therefore, finding the $f$ that precisely models the true generalizable correlation between $\bx$ and $y$ is challenging, especially with real-world data like natural images, which are complex and multiscale. In such scenarios, a neural network may compute incorrect (according to the ground-truth relationship between variables) yet perfect (in the empirical data distribution) correlations that explain the relationship between $\bx$ and $y$~\citep{goodfellow2016deep, neyshabur2017exploring}. This phenomenon is particularly evident when $y$ is entirely noise-based and independent of $\bx$, but the neural network still fits $y$ to $\bx$ perfectly~\citep{neyshabur2017exploring, zhang2017understanding}. This process, often called brute-force memorization~\citep{arpit2017closer, chatterjee2018learning}, involves the network creating intricate computational strategies to encode all $(\bx,y)$ pairs in the samples. Consequently, correlations established in this way are spurious, originating from sampling noise in the data rather than ground-truth relationships.

But how do humans distinguish generalizable correlations from spurious ones? Instead of relying on brute-force memorization to establish input-output correspondences, humans naturally focus on understanding high-level concepts within the input data, selectively ignoring irrelevant details~\citep{tomasello1999human, chopra2019first}. This approach leads to the formation of correlations through simple, comprehensible patterns~\citep{arpit2017closer, chaabouni2020compositionality}. Empirical evidence in emergent languages also suggests that the more compositional a language is, the more learners will use it~\citep{chaabouni2020compositionality,li2019ease}.

We can, therefore, define the distinctions between generalizable and spurious correlations. First, generalizable correlations are simple and comprehensible, exhibiting lower Kolmogorov Complexity~\citep{li2008introduction, goldblum2023no, sutskever_generalization2023}. Second, while there is only one ground-truth correlation for a dataset, the number of spurious correlations can be massive. These two major distinctions lead to the following hypothesis.

\paragraph{Hypothesis:} Generalizable correlations should be more easily imitable by learners compared to spurious correlations. Specifically, assume $T_G$ and $T_S$ are two teacher models that capture the generalizable correlation and spurious correlation from a dataset, respectively. We have student learners $S_G$ and $S_S$ that separately imitate $T_G$ and $T_S$:
\begin{itemize}
    \item From an effectiveness perspective, the final training and test losses of learner $S_G$ after training are typically lower than those of learner $S_S$.
    \item From an efficiency perspective, during training, the test losses of learner $S_G$ decrease more rapidly than those of $S_S$.
\end{itemize}
This hypothesis emphasizes that generalizable correlations inherent in data are not only more interpretable but also more readily imitable. It suggests that the inherent simplicity and uniqueness of generalizable correlations make them more attainable and recognizable for learning algorithms, in contrast to the complex and abundant nature of spurious correlations derived from noise. In the following we present our novel approach.

\subsection{Learning from Teaching Regularization}\label{Section_3_2}

Building upon the Hypothesis, we propose that the ease of imitation of the teacher model by student models can serve as a proxy for the generalizability of learned representations. By measuring the `imitability' of the teacher model in the learning process, we can infer the generalizability of it. A teacher that is easier to imitate implies higher generalization. We then design a novel regularization approach that involves training a teacher model $T$ alongside student models $S$ to imitate $T$, subsequently measuring the imitability of the teacher during training. We maximize imitability by incorporating it as an additional loss during the training of the teacher $T$. This imitability loss is termed the Learning from Teaching regularizer (\ours regularizer). By doing so, $T$ is optimized to be a teacher that is easier to imitate and, thus, possesses superior generalization compared to models without the \ours regularizer. We refer to this class of regularization methods as `Learning from Teaching Regularization' (\ours). \ours aligns with the broader concept of regularization in machine learning, where the goal is to promote generalizable representations and prevent overfitting.

Although \ours can be applied to supervised, unsupervised, and reinforcement learning, we begin our discussion with supervised learning. We train a network $T_{\btheta}$, parameterized by $\btheta$, as the main model, which also serves as the teacher model. Additionally, we train a set of $K$ networks $S_i, i=1, 2, \cdots, K$, as the student models\footnote{For convenience, $S$ is referred to as a single student learner henceforth.}. The total set of parameters of the $K$ networks is denoted by $\bphi$. Given a training dataset $\mathcal{D}_t=\{(\bx_1, y_1), \cdots, (\bx_n, y_n)\}$, we train $T$ and $S$ to model $p(y|\bx)$, denoted as $p_t(y|\bx)$ and $p_s(y|\bx)$, respectively. Additionally, \ours includes a predefined imitability metric $\mu_{s,t}(\cdot)=\mu(S(\cdot),T(\cdot))$. Intuitively, $\mu_{s,t}$ measures the difference between $S$ and $T$'s predictions on the same input (occasionally denoted as $\mu$ henceforth for convenience). There are many possible choices for the metric $\mu$, such as the $L^2$ loss between the hidden representations of a specific layer. In our experiments, we choose $\mu(\bx)=\mu_{\text{KL}}(p_s(y|\bx)||p_t(y|\bx))$, which is the KL-divergence~\citep{KL-divergence}, to quantify the distribution similarity between $S$ and $T$.

We first train the teacher model. The objective function of the teacher combines the regular task loss with the additional \ours regularizer $R(\btheta)$ (defined in Equation~\ref{Formula_multiple_students_regularizer}). For example, in supervised learning, we can use the negative log-likelihood loss for the regular task loss, and the objective function can be written as:
\begin{equation}
    L_t(\btheta) = -\frac{1}{|\mathcal{D}_t|}\sum_{(\bx_i, y_i)\in\mathcal{D}_t} \log p_t(y_i|\bx_i) + R(\btheta),
\end{equation}
where $|\mathcal{D}_t|$ is the number of samples in the dataset $\mathcal{D}_t$.

To train the student networks and enhance information diversity, we require an independent unlabelled dataset, denoted as $\mathcal{D}_s= \{\bx_1, \cdots, \bx_m\}$. This dataset can be identical to $\mathcal{D}_t$, or generated either by a generative model trained on $\mathcal{D}_t$ or through alternative augmentation methods (e.g. synthetic data generation). This unlabelled dataset constitutes the environment for the student networks to follow the prediction of the teacher and, therefore, explores and generalizes beyond the original training data.

Specifically, the student networks' goal is to imitate the correlations acquired by the teacher network during the training process. The training loss for students can be written as:
\begin{equation}
   L_{s}(\bphi) = \frac{1}{|\mathcal{D}_s|}\sum_{\bx\in \mathcal{D}_s}\sum_{i=1}^{K} \mu_{s_{i},t}(\bx),
\end{equation}
where $|\mathcal{D}_s|$ is the number of samples in the unlabelled dataset $\mathcal{D}_s$.
The loss function $L_{s}$ encourages the student networks to learn from the teacher network by minimizing the difference between their predictions, as measured by the metric $\mu_{s,t}(\bx)$.

The feedback from all students $S_i$ constitutes the \ours regularizer:
\begin{equation}
    R(\btheta) = \frac{\alpha}{|\mathcal{D}_s|}\sum_{\bx\in \mathcal{D}_s} \sum_{i=1}^{K} \lambda_i \mu_{t,s_i}(\bx),
    \label{Formula_multiple_students_regularizer}
\end{equation}
where $\lambda_i \geq 0$ represents the coefficient weight of the $i$-th student $S_i$, with $\sum_{i=1}^{K} \lambda_i = 1$. The $\lambda_i$ can be either a learnable parameter or fixed, such as $\frac{1}{K}$. Essentially, the \ours regularizer measures the imitability of the teacher. The regularization coefficient $\alpha$ controls the trade-off between the original task learning objective of $T$ and the feedback from the students.

The detailed procedure of \ours for supervised and unsupervised learning is outlined in Algorithm~\ref{Algorithm_LoT}, and \ours regularization for RL (using PPO as an example) is outlined in Algorithm~\ref{Algorithm_RL}. The teacher $T$ and student $S_i$ networks are initialized differently to ensure they learn diverse features and representations. In both algorithms, the teacher and student networks iteratively learn from each other, with the students imitating the teacher's correlations and the teacher incorporating the students' feedback into the learning process.

\begin{algorithm}[ht]
   \caption{Learning from Teaching Regularization}
   \label{Algorithm_LoT}
\begin{algorithmic}[1]
   \State {\bfseries Input:} Dataset $\mathcal{D}_s, \mathcal{D}_t$, Regularization Coefficient $\alpha>0$, Student Steps Ratio $N>0$
   \State Initialize teacher network $T$ parameterized by $\btheta$ and student networks $S_i, i=1,2,\cdots K$, parameterized by $\bphi$.
   \Repeat
   \State Sample a batch of data $\mathcal{B}_t\subset \mathcal{D}_t, \mathcal{B}_s\subset \mathcal{D}_s$
  \State Compute $\tilde{R}(\btheta)=\frac{\alpha}{|\mathcal{B}_s|}\sum_{\bx\in \mathcal{B}_s}\sum_{i=1}^{K} \lambda_i\mu_{t,s_i}(\bx)$
   \State Compute $\tilde{L}_t(\btheta)=-\frac{1}{|\mathcal{B}_t|}\sum_{(\bx,y)\in \mathcal{B}_t} \log p_t(y|\bx)$ + $\tilde{R}(\btheta)$
   \State Update $\btheta$ using gradient $\nabla_{\btheta}\tilde{L}_t(\btheta)$
   \For{$i=1$ {\bfseries to} $N$}
   \State Sample $\mathcal{B}_s\subset \mathcal{D}_s$
   \State  Compute $\tilde{L}_s(\bphi)=\frac{1}{|\mathcal{B}_s|}\sum_{\bx\in \mathcal{B}_s}\sum_{i=1}^{K}\mu_{s_i,t}(\bx)$
   \State  Update student networks' parameters $\bphi$ using loss gradient $\nabla_{\bphi}\tilde{L}_s(\bphi)$
   \EndFor
   \Until{$T$ converges}
\end{algorithmic}
\end{algorithm}

\subsection{Discussion}

The works most related to \ours are knowledge distillation (KD)~\citep{hinton2015distilling, furlanello2018born} and ease-of-teaching~\citep{li2019ease, chaabouni2020compositionality} in emergent languages. However, \ours differs significantly from these approaches. In KD, a teacher model containing task-specific knowledge transmits this knowledge to a student model (often smaller than the teacher), with the primary focus on the student's performance post-distillation. Conversely, in \ours, both the teacher and student models may lack or possess different task-specific knowledge. Generalization is improved through joint training, incorporating additional signals from student feedback. In emergent languages, \citet{li2019ease} propose that structured language is easier to teach to other agents than less structured ones, achieving higher task success rates with less training. Additionally, \citet{chaabouni2020compositionality} identify a strong positive correlation between language transmission efficiency to new message receivers and the degree of compositionality (structuredness) of the language. In \ours, we focus on tasks distinct from emergent languages, finding that generalizable correlations are easier to imitate. Under our Hypothesis, we design a novel \ours regularizer and algorithm to enhance the generalization of deep neural networks, extending the ease-of-teaching concept to supervised, unsupervised, and reinforcement learning. In parallel work, \citet{ning2024can} proposes Learning by Teaching (LbT), which utilizes teacher and student models to generate answers as training samples for the teacher model. However, the regularization method in \ours is fundamentally distinct from that in \citet{ning2024can}.

\section{Experiments}\label{section_experiment}
We first validate our Hypothesis in Section~\ref{section4_1}. Subsequently, we assess the performance of \ours across several tasks: Atari games (Section~\ref{section4_2}), language modeling (Section~\ref{section4_3}), and image classification (Section~\ref{section4_4}). We compare \ours to a Teacher-only baseline, wherein the regularization coefficient $\alpha$ in $R(\btheta)$ is set to 0, thereby blocking the student feedback. Unless specified otherwise, we employ only one student model. Except for the Atari games where the student can learn from the offline samples of the teacher, we set $N=1$ to manage computation (we study the impact of $N$ in Section~\ref{ablation_study}). Moreover, we study the computational efficiency and effects of hyperparameters of \ours in Sections~\ref{section_cost_analysis} and~\ref{ablation_study}.

\subsection{Generalizable Correlations are Easier to Imitate than Spurious Correlations.}\label{section4_1}
\begin{figure}[t]
  \centering
  \includegraphics[width=0.98\linewidth]{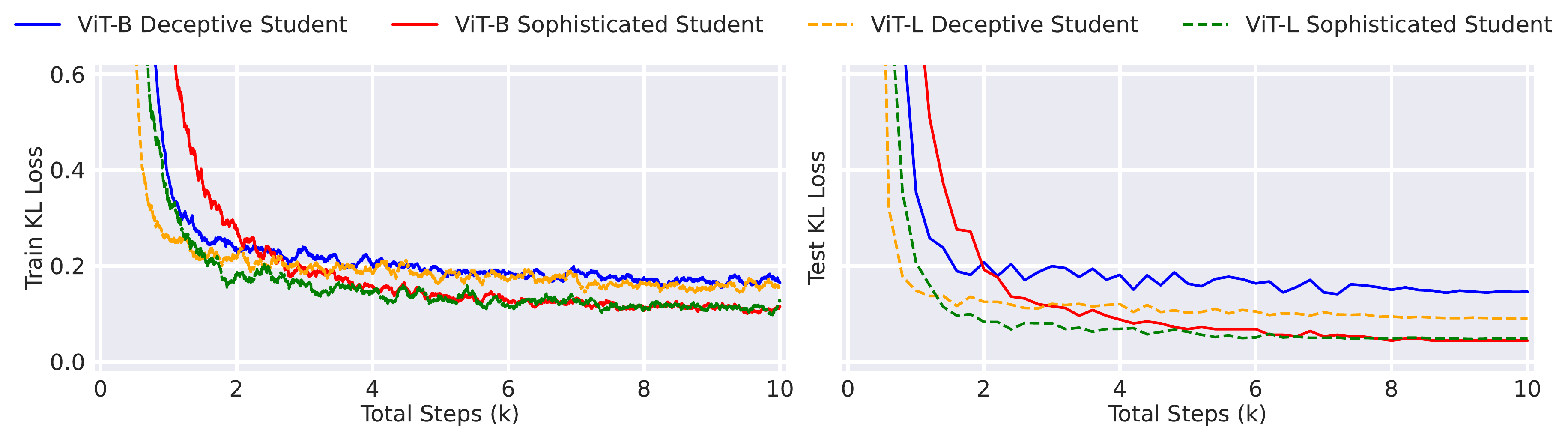}
  \caption{Training and test KL-divergence losses of student models in \ours using ViT-B/16 and ViT-L/16 on CIFAR-100 with different teacher models. The sophisticated students achieve lower losses than the deceptive students given the same computational budget.}
  \label{Figure_synthetic}
\end{figure}
In our Hypothesis, learners are presumed to more readily imitate generalizable correlations than spurious ones. To investigate this, we design experiments involving two distinct teacher models: a sophisticated teacher and a deceptive teacher. The sophisticated teacher effectively captures generalizable correlations, while the deceptive teacher primarily learns spurious correlations. We use an identical student model to learn from both teachers separately, monitoring the student-teacher KL divergence during training and testing. The student that learns easier-to-imitate correlations is expected to exhibit lower training and test KL losses with fewer training steps.

We employ the ViT-B/16 and ViT-L/16 architectures~\citep{dosovitskiy2020image} for both the teachers and students. The sophisticated teachers are trained on the full CIFAR-100~\citep{krizhevsky2009learning} training set for $10{,}000$ steps to achieve optimal convergence. The deceptive teachers, using the same hyperparameters and training steps as the sophisticated teachers, are trained on a random subset of $2,560$ images from the CIFAR-100 training set, leading to over-fitting. Consequently, the sophisticated teachers are expected to exhibit better generalization ability (their test accuracy surpasses that of the deceptive teachers by 14\%). 

The two student models referred to as the sophisticated student and the deceptive student, share identical hyperparameters and initializations. They are trained to imitate the correlations from their respective teachers on the full CIFAR-100 training set. The teacher models are kept frozen during the training of the students, with the objective $L_s(\bphi)$ defined as follows:
\begin{equation}
   L_s(\bphi) = \frac{1}{|\mathcal{D}_s|}\sum_{\bx \in \mathcal{D}_s} \mu_{\text{KL}}(p_s(y|\bx)||p_t(y|\bx)),
\end{equation}
where $\mathcal{D}_s$ represents the full training set of CIFAR-100.

We present the training and test losses in Figure~\ref{Figure_synthetic} and make the following observations:

\begin{itemize}
    \item Given the same computational budget, the sophisticated students achieve lower final KL losses on both the training and test sets compared to the deceptive students. This suggests that the student can more effectively imitate the prediction distribution of a teacher that captures generalizable correlations.
    \item The deceptive students require more training steps to achieve the same training and test student-teacher KL losses as the sophisticated students. This indicates that learners tend to grasp spurious correlations much more slowly than generalizable correlations.
\end{itemize}

These results suggest that generalizable correlations are easier to imitate than spurious ones. In \ours, we expect the teacher model to master generalizable correlations by incorporating feedback from students via the \ours regularizer.

\subsection{Atari Games}\label{section4_2}
We conduct experiments on four Atari games, namely BeamRider, Breakout, UpNDown, and Gravitar, following the implementation in~\citet{huang2022cleanrl}. Both the \ours and Teacher-only agents have identical hyperparameters. All agents are trained using Proximal Policy Optimization (PPO)~\citep{schulman2017proximal}. While the teacher agents interact with the game environment, the student agents are trained on the \textbf{most recent $10{,}240$ samples} generated by the teacher agents, ensuring that \ours and Teacher-only experience the same environmental interactions. We use different $\alpha$ values for various games and set $N=5$ to efficiently imitate the teacher. More details are provided in Appendix~\ref{Appendix_implementation_details}.

\begin{figure}[t]
  \centering
  \begin{minipage}[b]{0.24\linewidth}
    \includegraphics[width=\linewidth]{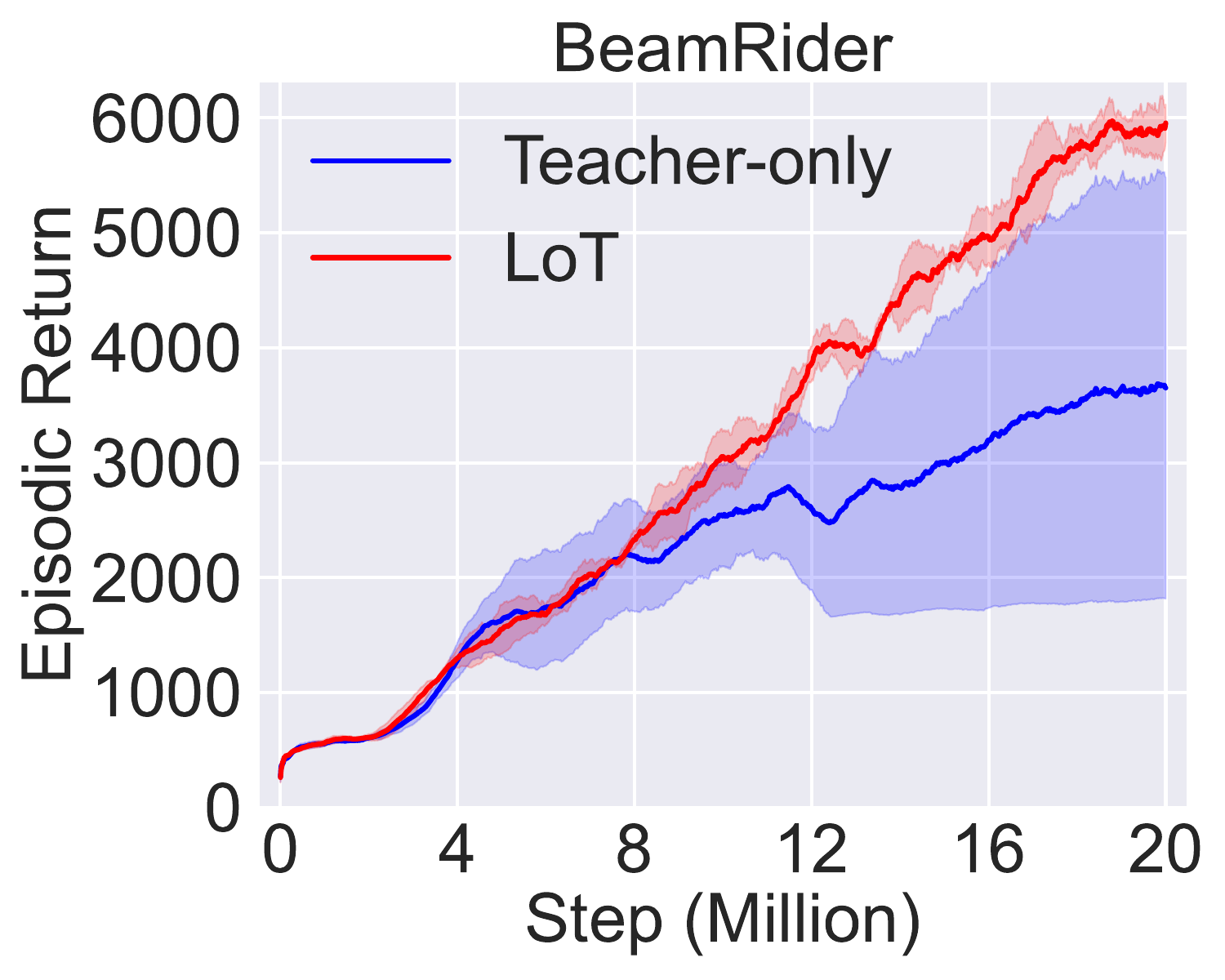}
  \end{minipage}
  \hfill
  \begin{minipage}[b]{0.24\linewidth}
    \includegraphics[width=\linewidth]{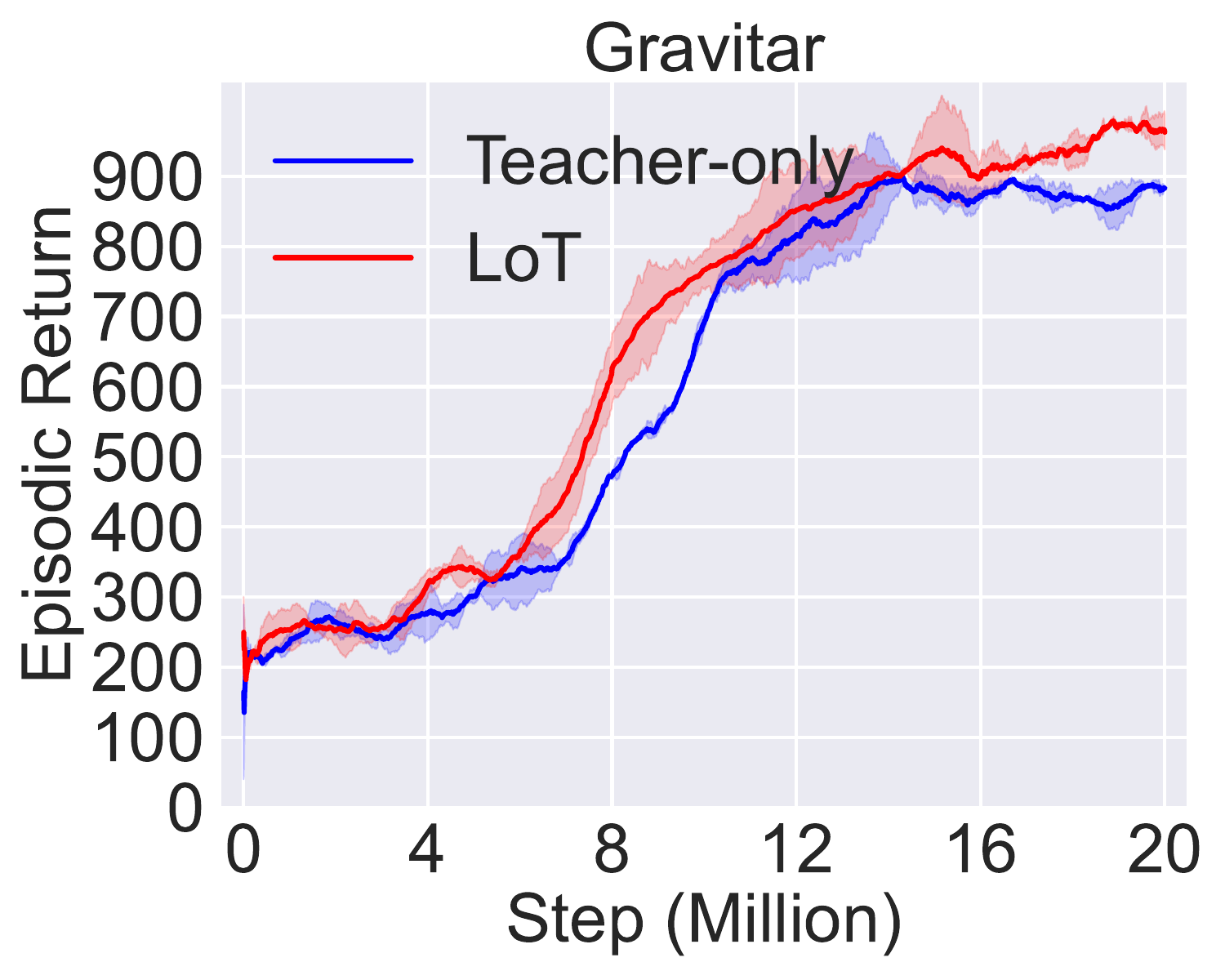}
  \end{minipage}
  \begin{minipage}[b]{0.24\linewidth}
    \includegraphics[width=\linewidth]{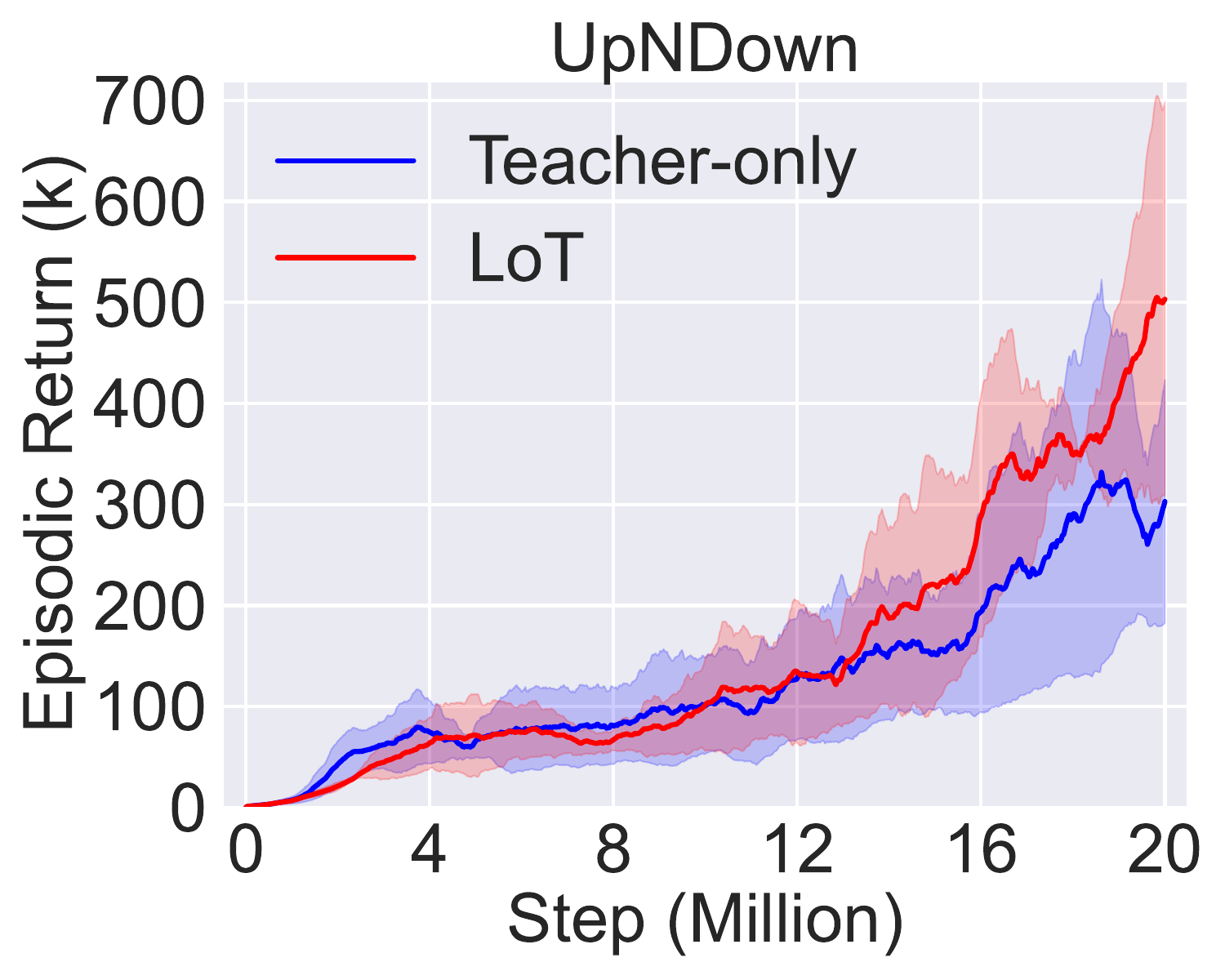}
  \end{minipage}\hfill
  \begin{minipage}[b]{0.24\linewidth}
    \includegraphics[width=\linewidth]{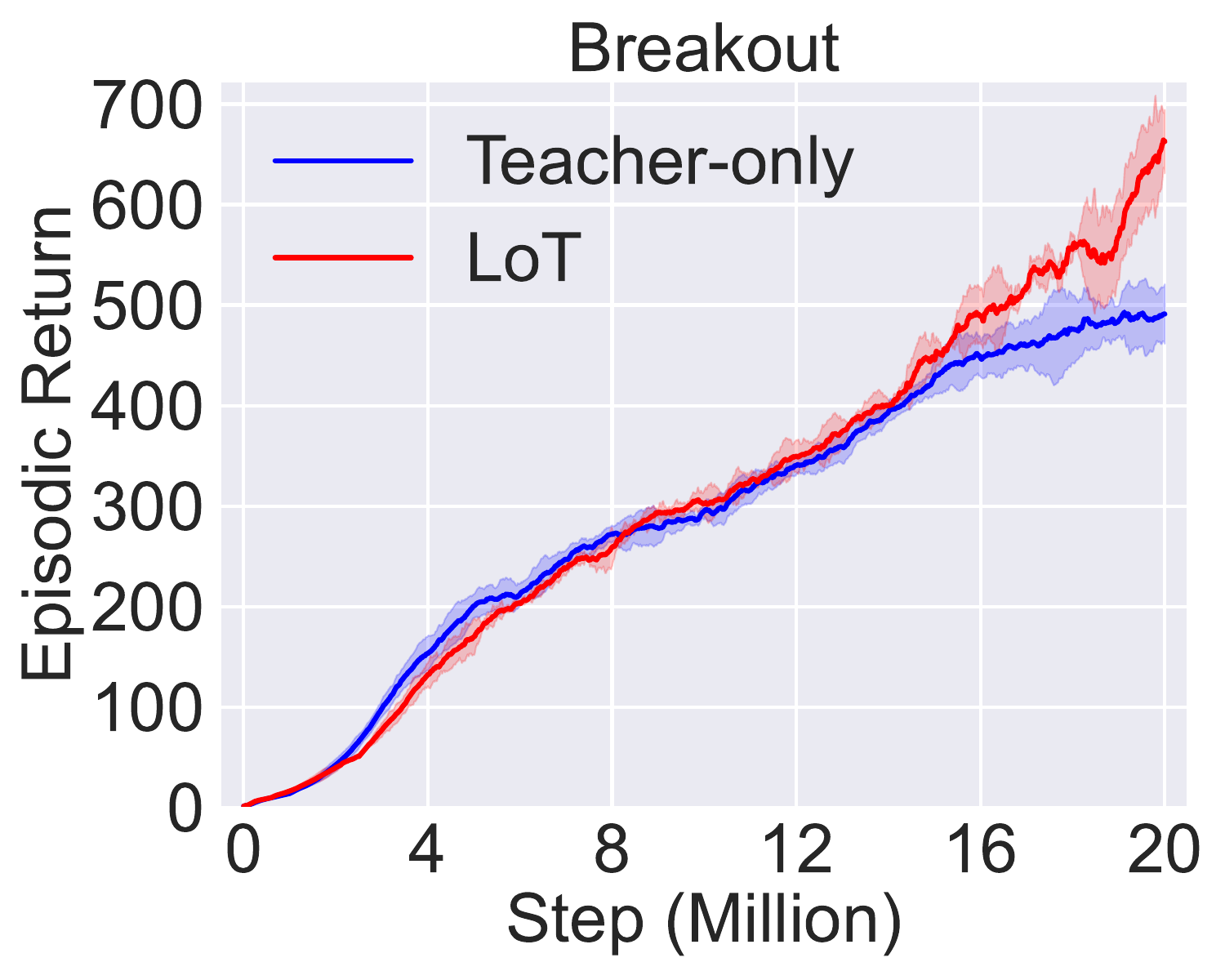}
  \end{minipage}
  \caption{The episodic return of 
the teacher agent in \ours and the Teacher-only on four Atari games (averaged over ten runs). \ours demonstrates return gains over Teacher-only on all games.}
  \label{Figure_RL_main_results}
\end{figure}

The empirical results are presented in Figure~\ref{Figure_RL_main_results}, and we make the following observations:
\begin{itemize}
    \item \ours improves the agent return compared to the Teacher-only version with $20$ million teacher training steps. Specifically, \ours achieves \{$63.14\%$, $9.79\%$, $66.48\%$, $35.70\%$\} normalized return enhancements on \{BeamRider, Gravitar, UpNDown, Breakout\}.
    \item The performance gain of \ours becomes more prominent as the training progresses (from $15$ million to $20$ million steps).
\end{itemize}
These results suggest that \ours is an effective approach for enhancing the generalization of RL agents, as it requires no additional environmental interactions while delivering significant performance gains.

\subsection{Language Modeling}\label{section4_3}
 
Language modeling is a widely acknowledged NLP task, and regularization techniques have been demonstrated to significantly enhance performance in this domain~\citep{zaremba2014recurrent}. To examine the impact of \ours on language modeling, we conduct experiments in two scenarios: unsupervised language pretraining and supervised fine-tuning.

\subsubsection{Unsupervised Language Pretraining}

We conduct experiments of \ours and Teacher-only using LSTM~\citep{hochreiter1997long}, AWD-LSTM~\citep{merity2018regularizing}, and Transformer-XL~\citep{dai2019transformer} for teacher and student on Penn Tree Bank (PTB)~\citep{marcus1993building} and WikiText-103~\citep{merity2017pointer}. 
We follow the implementations outlined in \citet{zaremba2014recurrent, merity2018regularizing, dai2019transformer}. In \ours, we utilize different coefficients $\alpha$ for various architectures and benchmarks to control the \ours regularizer. To ensure a fair comparison, we maintain the same total number of training steps (with teacher and student training steps accumulated) for \ours and the Teacher-only setup. Please refer to Appendix~\ref{Appendix_implementation_details} for more implementation details.

\begin{table}[t]
    \centering
    \caption{The test perplexity of the teacher model in \ours and the baseline on PTB and WikiText-103. Results are averaged over three runs. \ours achieves consistent perplexity reduction over different choices of architectures and benchmarks.}
    \label{Table_perf_language_modeling}
    \begin{center}
    \resizebox{0.95\textwidth}{!}{
    \begin{tabular}{c|cc|ccc}
    \toprule
    \textbf{Dataset} & \textbf{Teacher} & \textbf{Student} & \textbf{Teacher \#Param.} & \textbf{Teacher-only} & \textbf{\ours} \\ 
    \midrule
    \multirow{2}{*}{PTB} & LSTM & LSTM & 20M & $82.75 \pm 0.36$ & $\textbf{71.72} \pm 0.54$ \\
     & AWD-LSTM & AWD-LSTM & 24M & $58.69 \pm 0.37$ & $\textbf{53.31} \pm 0.56$ \\
    \midrule
    \multirow{2}{*}{WikiText-103} & Transformer-XL-B & Transformer-XL-B & 151M & $23.72 \pm 0.41$ & $\textbf{21.65} \pm 0.38$ \\
     & Transformer-XL-L & Transformer-XL-L & 257M & $18.50 \pm 0.25$ & $\textbf{16.47} \pm 0.23$ \\
    \bottomrule
    \end{tabular}}
    \end{center}
\end{table}

From the empirical results presented in Table~\ref{Table_perf_language_modeling}, we observe that \ours achieves notable perplexity (PPL) gains across various architectures and benchmarks under the same number of learning steps as Teacher-only. Specifically, \ours achieves at least 2 points PPL gains across all settings, and a $11.03$ gain for LSTM on PTB. It indicates that \ours can be effectively applied to both LSTM and Transformer architectures in language pretraining.

\subsubsection{Supervised Fine-tuning}

Furthermore, to evaluate the effectiveness of \ours in fine-tuning pretrained large language models (LLMs), we conduct supervised fine-tuning (SFT) experiments using 
LLaMA-1~\citep{touvron2023llama} and LLaMA-2~\citep{touvron2023llama2} on two mathematical reasoning benchmarks: GSM8K~\citep{cobbe2021training} and MATH~\citep{hendrycks2021measuring}.

We compare \ours to in-context learning (ICL)~\citep{brown2020language} and SFT. Following \citet{touvron2023llama2}, the number of in-context examples is 8 for GSM8K and 4 for MATH. The SFT configuration follows \citet{yue2023mammoth}, and we fine-tune the LLaMA models for four epochs. In \ours, the teacher and student models share the same architecture for simplicity. The models are trained for two epochs in \ours to match the total training steps in SFT for fair comparison. All other configurations are consistent with those used in SFT. More implementation details are described in Appendix~\ref{Appendix_implementation_details}.

\begin{wraptable}{r}{0.5\textwidth} 
    \vspace{-7mm}
    \centering
    \caption{The accuracy of the teacher model in \ours and the baseline on GSM8K and MATH. Results are averaged over three runs.}
    \vspace{1em}
    \label{Table_perf_language_modeling_finetune}
    \resizebox{0.5\textwidth}{!}{
    \begin{tabular}{c|cc}
    \toprule
    \textbf{Setting} & \textbf{GSM8K} & \textbf{MATH} \\ 
    \midrule
    $\text{LLaMA-1 7B}_{+ \text{ICL}}$ & $10.69 \pm 0.87$ & $2.84 \pm 0.25$ \\
    $\text{LLaMA-1 7B}_{+ \text{SFT}}$ & $34.39 \pm 1.28$ & $4.78 \pm 0.23$ \\
    $\text{LLaMA-1 7B}_{+ \text{\ours}}$ & $\textbf{36.42} \pm 1.46$ & $\textbf{5.39} \pm 0.28$ \\
    \midrule
    $\text{LLaMA-2 7B}_{+ \text{ICL}}$ & $14.62 \pm 0.96$ & $2.46 \pm 0.25$ \\
    $\text{LLaMA-2 7B}_{+ \text{SFT}}$ & $39.81 \pm 1.34$ & $5.79 \pm 0.31$ \\
    $\text{LLaMA-2 7B}_{+ \text{\ours}}$ & $\textbf{41.87} \pm 1.62$ & $\textbf{6.28} \pm 0.22$ \\
    \bottomrule
    \end{tabular}}
    \vspace{-6mm}
\end{wraptable}

We measure the accuracy of greedy decoding results in Table~\ref{Table_perf_language_modeling_finetune}, and we observe that \ours enhances reasoning abilities on all architecture and dataset choices. This indicates the competence of \ours in improving the fine-tuning performance with a computational cost comparable to SFT.

\subsection{Image Classification}\label{section4_4}

To investigate the effects of \ours on computer vision tasks, we apply \ours to image classification by conducting experiments using ResNets~\citep{he2016deep}, MobileNetV2~\citep{sandler2018mobilenetv2}, ViT~\citep{dosovitskiy2020image}, and Swin~\citep{liu2021swin} architectures pretrained on ImageNet-1K and ImageNet-21K~\citep{deng2009imagenet} as teacher and student models. We choose CIFAR-100~\citep{krizhevsky2009learning} and ImageNet-1K as the downstream datasets. The total training steps for \ours and the Teacher-only approach are the same for a fair comparison. Further implementation details are provided in Appendix~\ref{Appendix_implementation_details}. We conclude the following observations from results in Table~\ref{Table_perf_image_classification}:

\begin{table}[b]
\centering
\caption{The test accuracy of the teacher model for various teacher-student model combinations in \ours and the baseline. Results are averaged over three runs. \ours consistently enhances test performance in all model choices and datasets.\label{Table_perf_image_classification}}
\resizebox{1\textwidth}{!}{
\begin{tabular}{cc|cc|cccc}
\toprule
\textbf{Pretrained} & \textbf{Downstream} & \textbf{Teacher} & \textbf{Student} & \textbf{Image Size} & \textbf{Teacher/Student \#Param.} & \textbf{Teacher-only} & \textbf{\ours} \\ 
\midrule
\multirow{6}{*}{ImageNet-1K} & \multirow{6}{*}{CIFAR-100} & ResNet-18 & MobileNetV2 & $224^{2}$ & 12M / 4M & $81.14 \pm 0.58$ & $\textbf{82.78} \pm 0.36$  \\
 & & ResNet-18 & ResNet-18 & $224^{2}$ & 12M / 12M & $81.14 \pm 0.58$ & $\textbf{82.89} \pm 0.25$  \\
 & & ResNet-18 & ResNet-50 & $224^{2}$ & 12M / 26M & $81.14 \pm 0.58$ & $\textbf{83.13} \pm 0.26$  \\
 & & ResNet-50 & MobileNetV2 & $224^{2}$ & 26M / 4M & $84.09 \pm 0.32$ & $\textbf{85.38} \pm 0.44$  \\
 & & ResNet-50 & ResNet-18 & $224^{2}$ & 26M / 12M & $84.09 \pm 0.32$ & $\textbf{85.77} \pm 0.19$  \\
 & & ResNet-50 & ResNet-50 & $224^{2}$ & 26M / 26M & $84.09 \pm 0.32$ & $\textbf{86.04} \pm 0.38$  \\
\midrule
\multirow{4}{*}{ImageNet-21K} & \multirow{4}{*}{CIFAR-100} & ViT-B/16 & ViT-B/16 & $384^{2}$ & 86M / 86M & $91.57 \pm 0.31$ & $\textbf{93.17} \pm 0.35$ \\
 & & ViT-B/16  & ViT-L/16 & $384^{2}$ & 86M / 307M & $91.57 \pm 0.31$ & $\textbf{93.25} \pm 0.44$  \\
 & & ViT-L/16 & ViT-B/16 & $384^{2}$ & 307M / 86M & $93.44 \pm 0.28$ & $\textbf{94.29} \pm 0.33$  \\
 & & ViT-L/16 & ViT-L/16 & $384^{2}$ & 307M / 307M & $93.44 \pm 0.28$ & $\textbf{94.18} \pm 0.26$  \\
\midrule
\multirow{8}{*}{ImageNet-21K} & \multirow{8}{*}{ImageNet-1K} & ViT-B/16 & ViT-B/16 & $384^{2}$ & 86M / 86M & $83.97 \pm 0.11$ & $\textbf{84.54} \pm 0.15$ \\
 & & ViT-B/16  & ViT-L/16 & $384^{2}$ & 86M / 307M & $83.97 \pm 0.11$ & $\textbf{84.80} \pm 0.08$  \\
 & & ViT-L/16 & ViT-B/16 & $384^{2}$ & 307M / 86M & $85.15 \pm 0.17$ & $\textbf{85.92} \pm 0.09$  \\
 & & ViT-L/16 & ViT-L/16 & $384^{2}$ & 307M / 307M & $85.15 \pm 0.17$ & $\textbf{85.65} \pm 0.11$  \\
 & & Swin-B & Swin-B & $384^{2}$ & 88M / 88M & $86.37 \pm 0.06$ & $\textbf{86.68} \pm 0.15$ \\
 & & Swin-B & Swin-L & $384^{2}$ & 88M / 197M & $86.37 \pm 0.06$ & $\textbf{86.73} \pm 0.14$  \\
 & & Swin-L & Swin-B & $384^{2}$ & 197M / 88M & $87.27 \pm 0.11$ & $\textbf{87.64} \pm 0.12$  \\
 & & Swin-L & Swin-L & $384^{2}$ & 197M / 197M & $87.27 \pm 0.11$ & $\textbf{87.59} \pm 0.09$  \\
\bottomrule
\end{tabular}}
\end{table}

\begin{itemize}
    \item \ours achieves accuracy gains across various architectures and datasets without additional computational costs. For example, \ours improves test accuracy by almost 2 points using a ResNet-18 teacher and a ResNet-50 student on CIFAR-100 after pretrained on ImageNet-1K. Similarly, on the larger-scaled ImageNet dataset ImageNet-21K, \ours still obtains nearly 1 point improvement using ViT-B/16 as the teacher and ViT-L/16 as the student.
    \item The generalization of teacher models can be effectively enhanced by students of larger sizes. For instance, ResNet-50, ViT-L/16, and Swin-L students can enhance the performance of ResNet-18, ViT-B/16, and Swin-B teachers, respectively. Similarly, small student models can also enhance the generalization performance of larger teacher models using \ours. For example, a MobileNetV2 student improves the performance of RestNet-18 and ResNet-50 by more than 1 point on CIFAR-100 with a much smaller model size. Similar results appear on the ViT-L/16 teacher and ViT-B/16 student combination in the ImageNet-1K task.
    \item For transformer-based models, employing different architectures for teachers and students achieves better performance than sharing the same architecture. For example, when applying a ViT-B/16 student, a ViT-L/16 teacher achieves $0.27\%$ more accuracy than using a ViT-L/16 student. This suggests that using different architectures for teacher and student increases information diversity, which contributes to enhanced generalization for teacher models~\citep{sennrich2016improving}. 
\end{itemize}

These experimental results demonstrate the effectiveness of \ours in enhancing the generalization of pretrained CNN-based and Transformer-based vision models in image classification.

\subsection{Analysis of Computational Cost and Efficiency}\label{section_cost_analysis}

\begin{wrapfigure}{r}{0.45\textwidth}
  \vspace{-5mm}
  \centering
  \includegraphics[width=0.95\linewidth]{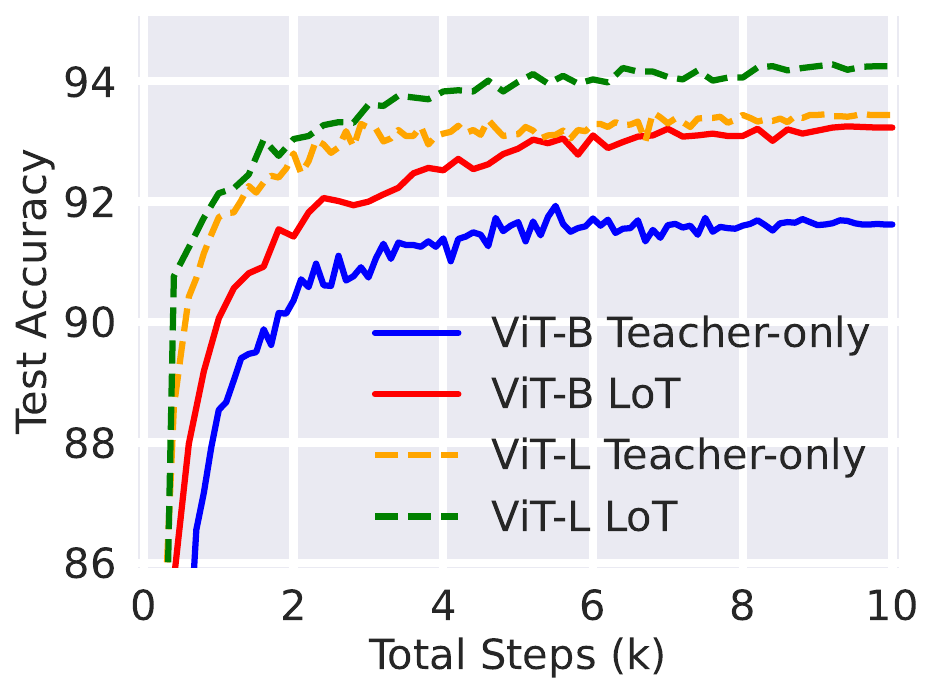}
  \caption{Test accuracy of teacher models in \ours and Teacher-only using ViT-B/16 and ViT-L/16 on CIFAR-100. \ours achieves higher test accuracy with fewer training steps.}
  \label{Figure_regularization}
\end{wrapfigure}

For supervised and unsupervised tasks, \ours involves training teacher models alongside student models as outlined in Algorithm~\ref{Algorithm_LoT}. Compared to Teacher-only, the potential limitation of \ours is that it requires additional computation and memory for the student models. Therefore, in our results in Section~\ref{section_experiment}, we maintain the same total training steps between \ours (accumulated for the teacher and student) and Teacher-only and demonstrate that \ours achieves better generalization performance under the same number of updates. In this regard, we show the test accuracy of image classification between \ours and Teacher-only using ViT models with respect to the total training steps in Figure~\ref{Figure_regularization}. We note that \ours achieves better test accuracy than Teacher-only in both ViT-B/16 and ViT-L/16 with fewer total training steps. Moreover, we demonstrate that \ours remains effective even when the student model is smaller than the teacher model in Table~\ref{Table_perf_image_classification}, which further reduces the computation cost compared to Teacher-only in the same total training steps and accommodates different student model choices with resource constraints. We provide more results regards efficiency of \ours in  Appendix~\ref{appendix_additional_results}.

In RL tasks, only the teacher model interacts with the environment to collect samples, and the student can learn from the teacher samples exclusively (please refer to Appendix~\ref{appendix_algorithm} for the algorithm of PPO-version \ours). Therefore, \ours introduces negligible computation costs since sample collections are more resource-intensive than fitting the agent network to the samples in RL. For instance, in our Atari games experiments, the training time of \ours (606 minutes) is comparable to the Teacher-only setting (597 minutes) on a single NVIDIA A6000 GPU.

\subsection{Additional Investigation}\label{ablation_study}

\paragraph{Comparison to KD.} To investigate the effect of \ours compared to other student-teacher learning paradigms, we compare \ours to the born-again networks (BAN) baseline~\citep{furlanello2018born}. In BAN, we select the checkpoint with the best performance of the Teacher-only model as the (frozen) teacher and distill its knowledge into a student model with an identical architecture. Equal weights are assigned to the hard loss (from the dataset) and soft loss (from the teacher) to train the student model~\citep{hinton2015distilling}. All other configurations remain consistent with \ours. The results in Table~\ref{Table_comparison_to_kd} indicate that \ours achieves superior performance than BAN with a strong feedback model, further indicating the significance of the interactive learning process in \ours.

\begin{table}[t]
\vspace{-1mm}
\centering
\caption{Performance comparison of Teacher-only, BAN and \ours on CIFAR-100. \ours achieves superior performance to Teacher-only and BAN.}
\label{Table_comparison_to_kd}
\resizebox{0.85\textwidth}{!}{
\begin{tabular}{c|cc|ccc}
\toprule
\textbf{Dataset} & \textbf{Teacher} & \textbf{Student} & \textbf{Teacher-only} & \textbf{BAN (Student)} & \textbf{\ours (Teacher)} \\ 
\midrule
CIFAR-100 & ResNet-18 & ResNet-18 & 81.14 & 82.08 & \textbf{82.89} \\
CIFAR-100 & ResNet-50 & ResNet-50 & 84.09 & 84.73 & \textbf{86.04} \\
CIFAR-100 & ViT-B/16 & ViT-B/16 & 91.57 & 92.44 & \textbf{93.17} \\
CIFAR-100 & ViT-L/16 & ViT-L/16 & 93.44 & 93.82 & \textbf{94.18} \\
\bottomrule
\end{tabular}}
\end{table}

\begin{figure}[t]
\vspace{-1mm}
  \centering
   \includegraphics[width=0.95\linewidth]{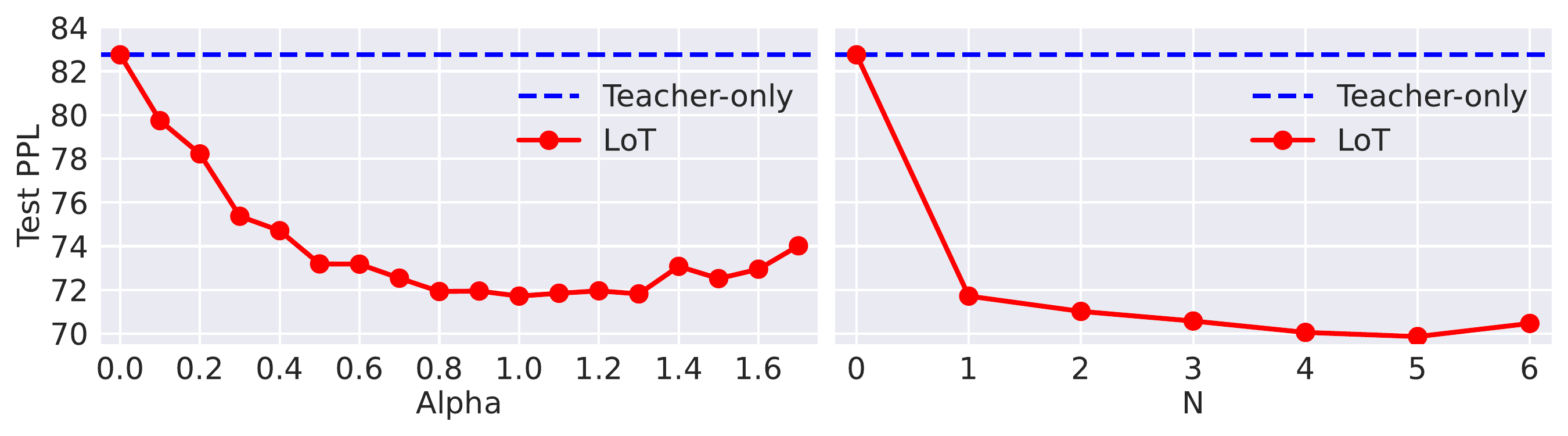}
  \caption{Effects of regularization coefficient $\alpha$ (left) and student steps ratio $N$ (right). $\alpha=1$ is the best $\alpha$ value to achieve the lowest test perplexity of the teacher model, and moderate student steps ratio $N$ such as 4 and 5 benefit the teacher model the most.}
  \label{Figure_Ablation_Alpha_Student_Ratio}
\end{figure}

\paragraph{Effect of regularization coefficient $\alpha$.}
The strength of regularization plays a crucial role in the overall training effect~\citep{krizhevsky2012imagenet}. To investigate the effects of \ours on the generalization of the teacher model, we perform experiments on PTB using the LSTM architecture for both teacher and student models. The configuration follows Section~\ref{section4_3}, except that we gradually increase the value of $\alpha$ in \ours from $0$ to $1.7$ and examine the test PPL of the teacher model. The results are presented in Figure~\ref{Figure_Ablation_Alpha_Student_Ratio} (left). We observe that the performance of the teacher model improves rapidly as $\alpha$ increases from $0$ to $1$, and when the value exceeds this point, the performance of the teacher begins to decline. This observation suggests that moderate feedback from the student is most beneficial for the teacher, but an excessively strong signal can hinder the teacher's learning process. Similar effects of large $\alpha$ values have been noted in joint teacher-student training in knowledge distillation~\cite{park2021learning}. 

\paragraph{Effect of student steps ratio $N$.}
To demonstrate the importance of the student steps ratio $N$ in \ours, we conduct additional experiments by training LSTM teacher and student models on PTB using various values of $N$. The empirical results presented in Figure~\ref{Figure_Ablation_Alpha_Student_Ratio} (right) indicate that the teacher benefits most from a moderate $N$ value, such as $4$ or $5$. This finding suggests that achieving a balanced ratio between teacher and student model updates is crucial for optimal performance. When $N$ is too low, the student may not sufficiently learn from the teacher, thereby reducing the quality of the feedback it provides. Conversely, if $N$ is too high, the student may overfit the teacher's errors, resulting in less effective imitability measurement.

\section{Conclusion}

Identifying generalizable multiscale correlations from the vast space of possible correlations remains a significant challenge in machine learning. Inspired by cognitive science beliefs about human intelligence, we have shown experimentally that generalizable correlations are more imitable by other learners. In particular, we introduced a novel regularization method, \ours, which identifies generalizable correlations by teaching student models and exploiting their feedback. We conducted comprehensive experiments across various learning tasks and neural architectures. The results demonstrate that our proposed regularizer enhances model performance effectively and efficiently. In conclusion, our proposed \ours regularization offers a promising new approach to improve the generalization of neural networks by leveraging the learning process of student models and incorporating their feedback to refine the teacher model.

\section{Acknowledgments}
Metaxas is partially supported by research grants from NSF: 2310966, 2235405, 2212301, 2003874, 1951890, AFOSR 23RT0630, and NIH 2R01HL127661.

\bibliography{neurips_2024}

\newpage
\appendix
\section{Ethics and Social Impacts}\label{appendix_social_impacts}
In this work, we propose a regularization method to improve the generalization of deep neural networks. Our work focuses on technical contributions to deep learning and AI. Therefore, the potential social impacts of AI in general apply to our work, including fake information, toxic content, fairness concerns, and misuse of AI. For example, toxic content like hate speech can lead to data contamination and therefore have harmful impacts on society, which has been observed in large-scale pretrained models. By employing our method, such harmful behavior can potentially be amplified. 

\section{Related Works}\label{appendix_related_works}
\subsection{Regularization in Deep Learning}
Regularization serves as a primary strategy to improve generalization capabilities and mitigate over-fitting~\citep{labach2019survey}. Various regularization techniques exist for deep neural networks. One of the earliest and most straightforward approaches to regularization involves constraining the model's capacity by adding a penalty function to the original objective function. Techniques such as L1 regularization~\citep{hoerl1970ridge, tibshirani1996regression, tibshirani1997lasso}, L2 regularization~\citep{saunders1998ridge, cortes2012l2, krizhevsky2012imagenet}, and weight decay~\citep{yu2008deep, krizhevsky2012imagenet} fall into this category. Introducing noise~\citep{poole2014analyzing,hochreiter1994simplifying} to the system can also judiciously enhance generalizability and prevent over-fitting. Dropout~\citep{hinton2012improving, wan2013regularization, ba2013adaptive, srivastava2014dropout} is a widely used regularization technique that randomly drops certain neural network connections during training.

\subsection{Student-Teacher Learning Paradigms}
\subsubsection{Knowledge Distillation}

Knowledge distillation (KD) is a technique that transfers knowledge from a teacher model to a student model by training the student to imitate the teacher's outputs~\citep{hinton2015distilling}. This approach is widely applied in areas such as model compression, transparency, and interpretability~\citep{buciluǎ2006model, hinton2015distilling, breiman1996born, tan2018transparent, frosst2017distilling, lipton2016mythos}. Model compression is often motivated by resource constraints. Pioneering works include \citet{buciluǎ2006model}, which compresses ensemble neural networks into a single network, and \citet{ba2014deep}, which improves shallow neural network accuracy by mimicking deep networks. KD is also applied in various domains, including deep reinforcement learning~\citep{rusu2015policy}, continual learning~\citep{furlanello2016active, li2016learning, shin2017continual}, and learning privileged information theory~\citep{pechyony2010theory, lopez2015unifying}. The dark knowledge method~\citep{hinton2015distilling} further develops KD, where a student model aims to fully match the output distribution of the teacher. Intuitively, distillation is effective because the teacher's output distribution over classes provides a more informative training signal than a one-hot label. Additionally, in born-again networks (BAN)~\citep{furlanello2018born}, the teacher and student have identical neural architecture and model sizes, but the student can surprisingly surpass the teacher's accuracy.

\subsubsection{Language Emergence}
In a cooperative environment, agents can learn emergent languages for communication to solve specific tasks. The emergence of such communication protocols is extensively studied in the context of multi-agent referential games~\citep{lewis2008convention, foerster2016learning}. In these games, one agent is required to describe its observations to another agent, which is then tasked with deducing the initial agent's observations~\citep{lazaridou2016multi}. The majority of methods employed to learn discrete communication protocols between agents utilize RL~\citep{foerster2016learning, vani2021iterated}. Compositionality is a desirable feature in the language used by agents, as it facilitates flawless generalization for previously unseen combinations of attributes~\citep{kottur2017natural, choi2018multi, bahdanau2018systematic, chaabouni2020compositionality}. However, the community still lacks strong research indicating what general conditions are necessary or sufficient for compositional language emergence. \citet{li2019ease, chaabouni2020compositionality, chaabouni2022emergent, galke2023makes} postulate that compositional languages are more straightforward to learn. 

\section{Motivation and Insights of Our Method}

The concept of Learning from Teaching originates in cognitive psychology and linguistics, particularly within the iterated learning theory of language emergence \citep{smith2003iterated,kirby2014iterated,kirby2008cumulative,kirby2001spontaneous}. This theory posits that the generalizable nature of languages arises from the iterative learning process across generations in a society. The core hypothesis is that a generalizable language is inherently easier to teach and learn \citep{li2019ease,vaniiterated,ren2024improving}, which aligns with our main hypothesis.

In the AI community, recent research has aimed to employ iterated learning to enhance the generalization of emergent languages and language acquisition among artificial learners. For example, some studies have used iterated learning to improve the generalization of emergent languages between AI agents \citep{li2019ease,ren2024improving}, while others have applied it to address generalization challenges in tasks like compositional Visual Question Answering (VQA) \citep{vaniiterated}. \ours shares the same motivation as this line of research. Our primary contribution extends the concept of “ease-of-teaching” \citep{li2019ease} from language learning to a broader range of machine learning tasks, including supervised, unsupervised learning, and reinforcement learning.

\ours functions as a regularizer, similar to other commonly used regularizers like the L2 regularizer. The L2 regularizer is effective because it encourages neural networks to learn simpler correlations, thereby avoiding overfitting. It is widely accepted that correlations with lower Kolmogorov complexity are more generalizable if they can perfectly explain a complex dataset. This aligns with the idea that "generalization equals optimal compression," as discussed by Ilya Sutskever \citep{sutskever2023observation}. Essentially, this notion adapts Occam’s Razor to the field of AI. Our key insight is that the "ease-of-teaching" metric serves as an effective regularizer beyond language emergence tasks.

Consider an intuitive example: Student A learns math by rote memorization, while Student B understands the core concepts and only memorizes essential rules, deducing the rest when needed. Both approaches can perform similarly on simple problem sets. However, as data complexity increases, Student A's burden grows significantly, while Student B's understanding-based approach remains manageable. Consequently, Student B's knowledge is easier to teach to another student, as it involves less complexity. Therefore, teachability (or imitability) can serve as a proxy for complexity.

\vspace{-1mm}
\section{Implementation Details.}\label{Appendix_implementation_details}
\vspace{-1mm}
\paragraph{Atari Games.} We perform experiments on four Atari games, namely Beam-Rider, Breakout, UpNDown, and Gravitar, following the implementation outlined in~\citep{huang2022cleanrl}. We set the regularization coefficient $\alpha$ to $0.5$ for BeamRider, Breakout, and UpNDown, and to $0.1$ for Gravitar. The other hyperparameters remain consistent across all four games. We use $N$ of $5$. For all agents, the optimizer employed is Adam, with an initial learning rate of $0.00025$. The teacher agent is trained for a total of $20{,}000{,}000$ timesteps. The temperature used in the KL loss is set to $1$. The experiments are implemented on the NVIDIA A6000 48GB GPUs. 

\paragraph{Language Modeling.} In the training-from-scratch experiments, we use the Transformer-XL architecture following \citet{dai2019transformer}, the LSTM architecture following \citet{zaremba2014recurrent}, and the AWD-LSTM architecture following \citet{merity2018regularizing}. For supervised fine-tuning experiments with LLaMA-1 and LLaMA-2, we employ the hyperparameters described in \citet{yue2023mammoth} and use the HuggingFace Transformers library~\citep{wolf2019huggingface}. The hyperparameters for \ours are detailed in Table~\ref{Table_Appendix_Language_Modeling}. The experiments for LSTM and AWD-LSTM are implemented on one single NVIDIA A100 40GB GPU. The Transformer-XL and LLaMA of \ours are trained on 4 and 8 NVIDIA A100 40GB GPUs, respectively.

\begin{table}[ht]
\centering
\begin{center}
\resizebox{0.95\columnwidth}{!}{
\begin{tabular}{ll|cccccc}
\toprule
Model & Dataset & $\alpha$ & $N$ & Optimizer & Learning Rate  & Training Epochs/Steps & Temperature  \\ 
\midrule
LSTM           & PTB          & $1.0$ & $1$ & SGD & $30$ & $30$ Epochs & $1.5$   \\
AWD-LSTM       & PTB          & $1.0$ & $1$ & ASGD & $30$ & $250$ Epochs & $1.5$ \\
Transformer-XL-B & WikiText-103 & $0.1$ & $1$ & ADAM & $0.01$ & $60{,}000$ Steps  & $2$   \\
Transformer-XL-L & WikiText-103 & $0.1$ & $1$ & ADAM & $0.01$ & $150{,}000$ Steps & $2$   \\
\midrule
LLaMA-1 7B     & GSM8K     & $0.01$ & $1$ & ADAMW & $ 2\times 10^{-5}$ & $2$ Epochs & $2$   \\
LLaMA-1 7B     & MATH      & $0.01$ & $1$ & ADAMW & $ 2\times 10^{-5}$ & $2$ Epochs & $2$   \\
LLaMA-2 7B     & GSM8K     & $0.01$ & $1$ & ADAMW & $ 2\times 10^{-5}$ & $2$ Epochs & $2$   \\
LLaMA-2 7B     & MATH      & $0.01$ & $1$ & ADAMW & $ 2\times 10^{-5}$ & $2$ Epochs & $2$   \\
\bottomrule
\end{tabular}}
\end{center}
\caption{Hyperparameters for Language Modeling.}
\label{Table_Appendix_Language_Modeling}
\vspace{-6mm}
\end{table}

\paragraph{Image Classification.} For CNN experiments, we use the ImageNet-1K pretrained architectures MobileNetV2 and ResNets, which can be downloaded from the official PyTorch Model Zoo\footnote{\url{https://pytorch.org/vision/stable/models.html}}. For ViT and Swin experiments, we follow the implementations described in \citet{dosovitskiy2020image} and \citet{liu2021swin}, using the official ImageNet-1K or ImageNet-21K pretrained weights downloaded from \footnote{\url{https://github.com/google-research/vision_transformer}} and \footnote{\url{https://github.com/microsoft/Swin-Transformer}}. The optimal hyperparameters for \ours are obtained through grid research. The detailed hyperparameters are illustrated in Table~\ref{Table_Appendix_Image_classification}. The experiments for MobileNetV2 and ResNets are implemented on one single NVIDIA A100 40GB GPU. The ViT and Swin experiments are implemented on 4 NVIDIA A100 40GB GPUs.

\begin{table}[ht]
\centering
\begin{center}
\resizebox{0.95\columnwidth}{!}{
\begin{tabular}{ll|cccccc}
\toprule
Model & Dataset & $\alpha$ & $N$ & Optimizer & Learning Rate  & Training Epochs/Steps & Temperature  \\ 
\midrule
MobileNetV2 & CIFAR-100 & $1.0$ & $1$ & SGD & $0.02$ & $30$ Epochs & $1.5$ \\
ResNet-18   & CIFAR-100 & $1.0$ & $1$ & SGD & $0.02$ & $30$ Epochs & $1.5$ \\
ResNet-50   & CIFAR-100 & $1.0$ & $1$ & SGD & $0.02$ & $30$ Epochs & $1.5$   \\
\midrule
ViT-B/16 & CIFAR-100 & $1.0$ & $1$ & SGD & $0.02$ & $5{,}000$ Steps & $1.5$   \\
ViT-L/16 & CIFAR-100 & $1.0$ & $1$ & SGD & $0.02$ & $5{,}000$ Steps & $1.5$   \\
ViT-B/16 & ImageNet-1K & $1.0$ & $1$ & SGD & $0.03$ & $10{,}000$ Steps & $1.5$   \\
ViT-L/16 & ImageNet-1K & $1.0$ & $1$ & SGD & $0.03$ & $10{,}000$ Steps & $1.5$   \\
\midrule
Swin-B   & ImageNet-1K  & $0.5$ & $1$ & ADAMW & $ 2\times 10^{-5}$ & $15$ Epochs & $1.5$   \\
Swin-L   & ImageNet-1K  & $0.5$ & $1$ & ADAMW & $ 2\times 10^{-5}$ & $15$ Epochs & $1.5$   \\
\bottomrule
\end{tabular}}
\end{center}
\caption{Hyperparameters for Image Classification.}
\label{Table_Appendix_Image_classification}
\vspace{-8mm}
\end{table}

\section{Scalability Analysis}\label{appendix_scalability_analysis}
\vspace{-2mm}
From our extensive results shown in Section~\ref{section_experiment}, \ours proves to be widely applicable across various domains, including reinforcement learning (Section~\ref{section4_2}), unsupervised learning (Section~\ref{section4_3}), and supervised learning (Section~\ref{section4_4}). It can be effectively applied to different architectures such as CNN-based (Table~\ref{Table_perf_image_classification}), LSTM-based (Table~\ref{Table_perf_language_modeling}), and Transformer-based(Table~\ref{Table_perf_language_modeling}) models. \ours works well on both small datasets like PTB (Table~\ref{Table_perf_language_modeling}) and CIFAR-100 (Table~\ref{Table_perf_image_classification}), and large datasets such as WikiText-103 (Table~\ref{Table_perf_language_modeling}) and ImageNet (Table~\ref{Table_perf_image_classification}). It is also suitable for both small models like ResNets (Table~\ref{Table_perf_image_classification}) and large models like ViT (Table~\ref{Table_perf_image_classification}) and LLaMA (Table~\ref{Table_perf_language_modeling_finetune}). Additionally, \ours is compatible with existing regularization methods such as weight decay and dropout. In our experiments with ResNets, weight decay was applied to both \ours and Teacher-only setups. In the experiments with Transformer-XL, ViT, and Swin, dropout is applied to both \ours and Teacher-only setups.
\vspace{-3mm}
\section{Limitation}\label{appendix_limitation}
\vspace{-2mm}
A potential limitation of \ours lies in the additional computational and memory costs required for training the student models. However, as demonstrated in Section~\ref{section_cost_analysis}, \ours achieves better generalization with fewer training steps compared to Teacher-only models, and the flexibility in choosing student models can accommodate varying resource constraints. In RL, the additional computational costs introduced by \ours are negligible, as sample collection is more resource-intensive than fitting the agent networks to the samples, as discussed in Section~\ref{section_cost_analysis}. Moreover, in real-world settings, inference cost is more critical than training cost. The superior generalization achieved by \ours offers significant benefits during inference without introducing additional inference costs.

\vspace{-2mm}
\section{Algorithm for the PPO-version of Our Method}\label{appendix_algorithm}
\vspace{-1mm}
The \ours algorithm for Proximal Policy Optimization (PPO) is illustrated in Algorithm~\ref{Algorithm_RL}. In our experiments, the teacher's sampled data $\mathcal{B}_t$ is continuously added to the student sample collections $\mathcal{D}_s$. Meanwhile, the most recent samples from $\mathcal{D}_s$ are used to formulate the student training batch $\mathcal{B}_s$ to ensure a high quality of its training dataset.

\vspace{-2mm}
\begin{algorithm}[ht]
   \caption{Learning from Teaching for PPO}
   \label{Algorithm_RL}
\begin{algorithmic}[1]
   \State {\bfseries Input:} Regularization Coefficient $\alpha>0$, Student Steps Ratio $N>0$.
   \State Initialize teacher network $T$ parameterized by $\btheta$ and student networks $S_i, i=1,2,\cdots K$, parameterized by $\bphi$.
   \State Initialize replay buffer $\mathcal{D}_s=\emptyset$
   \Repeat
   \State Sample minibatch $\mathcal{B}_t$ by running $T$ in simulator, add $\mathcal{B}_t$ to  $\mathcal{D}_s$
   \State Sample a batch of data $\mathcal{B}_s\subset \mathcal{D}_s$
   \State Compute $\tilde{R}(\btheta)=\frac{\alpha}{|\mathcal{B}_s|}\sum_{\bx\in \mathcal{B}_s}\sum_{i=1}^{K} \lambda_i\mu_{t,s_i}(\bx)$
   \State Compute $\tilde{L}_t(\btheta)$ using the PPO loss on minibatch $\mathcal{B}_t$
   \State Update $\btheta$ using gradient $\nabla_{\btheta}\tilde{L}_t(\btheta)$
   \State Fit value network for PPO on minibatch $\mathcal{B}_t$
   \For{$i=1$ {\bfseries to} $N$}
   \State Sample $\mathcal{B}_s\subset \mathcal{D}_s$
   \State  Compute $\tilde{L}_s(\bphi)=\frac{1}{|\mathcal{B}_s|}\sum_{\bx\in \mathcal{B}_s}\sum_{i=1}^{K}\mu_{s_i,t}(\bx)$
   \State  Update student networks' parameters $\bphi$ using loss gradient $\nabla_{\bphi}\tilde{L}_s(\bphi)$
   \EndFor
   \Until{$T$ converges}
\end{algorithmic}
\end{algorithm}
\vspace{-2mm}

\section{Additional Results}\label{appendix_additional_results}
\vspace{-2mm}
\paragraph{Computational Efficiency.} To further demonstrate the computational efficiency and superiority of \ours, we conduct experiments using LSTM on PTB and ViT-B/16 on CIFAR-100 with varying training epochs and steps, while keeping other configurations the same as in Section~\ref{section4_3} and Section~\ref{section4_4}. The results presented in Table~\ref{Table_multiple_training_steps} demonstrate that given equivalent computational budgets, \ours consistently outperforms the Teacher-only model across various datasets and architectures, even when the Teacher-only model trains for twice the number of epochs and steps. This further highlights \ours's effectiveness in improving the teacher model's generalization while maintaining enhanced computational efficiency.

\begin{table}[ht]
\centering
\caption{Performance of the teacher model in \ours and Teacher-only on image classification. The hyperparameters are the same as the corresponding experiments in the paper.}
\label{Table_multiple_training_steps}
\resizebox{0.8\textwidth}{!}{
\begin{tabular}{c|cc|ccc}
\toprule
\textbf{Dataset} & \textbf{Teacher} & \textbf{Student} & \textbf{Total Train Epochs/Steps} & \textbf{Teacher-only} & \textbf{\ours} \\
\midrule
CIFAR-100 & ViT-B/16 & ViT-B/16 & 10,000 steps & 91.57 & \textbf{93.17} \\
CIFAR-100 & ViT-B/16 & ViT-B/16 & 15,000 steps & 91.74 & \textbf{93.23} \\
CIFAR-100 & ViT-B/16 & ViT-B/16 & 20,000 steps & 91.82 & \textbf{93.40} \\
\midrule
PTB & LSTM & LSTM & 60 epochs & 82.75 & \textbf{71.72} \\
PTB & LSTM & LSTM & 90 epochs & 82.48 & \textbf{71.22} \\
PTB & LSTM & LSTM & 120 epochs & 82.42 & \textbf{70.67} \\
\bottomrule
\end{tabular}}
\end{table}

\begin{figure}[t]
  \centering
  \includegraphics[width=1\linewidth]{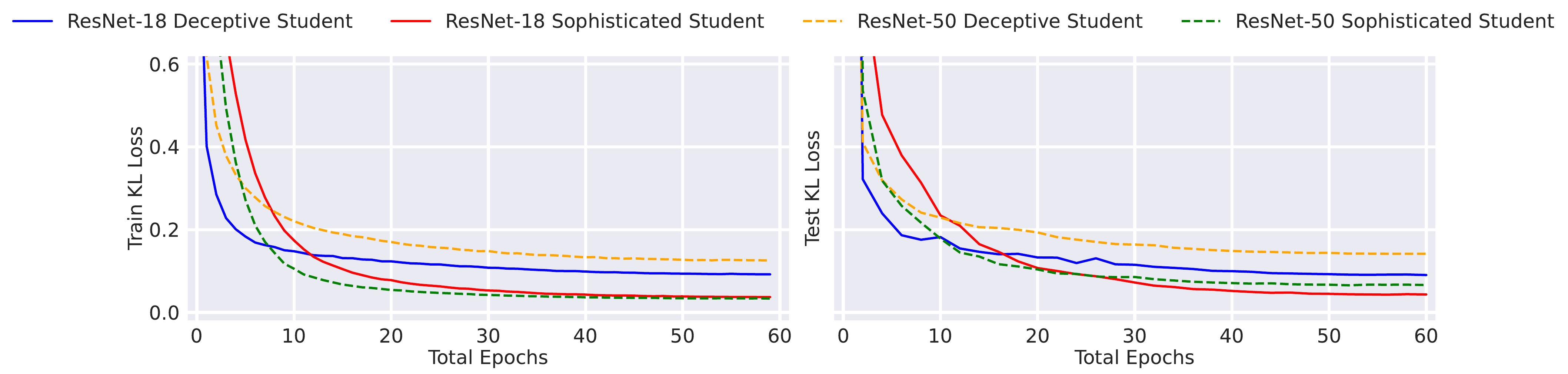}
  \caption{Training and test KL-divergence losses of student models in \ours using ResNet-18 and ResNet-50 on CIFAR-100 with different teacher models.}
  \label{Figure_synthetic_more}
\end{figure}

\paragraph{Additional Evidence for Hypothesis.} We provide additional experimental results to validate our hypothesis using ResNet-50 and ResNet-18 as both the teacher and student models on CIFAR-100, following the same methodology described in Section \ref{section4_1}, but with different model architectures. The training and test KL-divergence of the sophisticated and deceptive students are shown in Figure \ref{Figure_synthetic_more}. We observe that the sophisticated students achieve lower final KL losses compared to the deceptive students with fewer training epochs, which further supports our hypothesis

\paragraph{Out-of-distribution Performance.} We conduct additional experiments by fine-tuning models on ImageNet-1K and evaluating them on ImageNet-R and ImageNet-Sketch using ViT-B/16 and ViT-L/16 models to investigate the out-of-distribution robustness of \ours. The results, shown in Table \ref{Table_R7_performance_improvement}, demonstrate that \ours also brings performance improvements on these datasets, indicating the robustness of \ours across a broader set of scenarios.

\paragraph{Additional Comparison to KD.} In Table \ref{Table_comparison_to_kd}, we show that \ours outperforms the distillation method BAN. To provide stronger validation of \ours’s effectiveness, we conduct additional experiments using ResNet-50 and ViT-B/16 on CIFAR-100. We compare \ours to distillation methods such as BAN, DKD \citep{zhao2022decoupled}, and ReviewKD \citep{chen2021distilling}, with the teacher weights in these methods being the best checkpoint of Teacher-only. The results, shown in Table \ref{Table_performance_comparison}, indicate that \ours achieves better performance than these distillation baselines, further underscoring the effectiveness of the unique interactive learning process of \ours.

\begin{table}[ht]
\centering
\label{Table_combined}
\resizebox{\textwidth}{!}{
\begin{minipage}{0.52\textwidth}
    \centering
    \caption{Performance of \ours and Teacher-only on ImageNet-R and ImageNet-Sketch.}
    \label{Table_R7_performance_improvement}
    \resizebox{0.98\linewidth}{!}{
    \begin{tabular}{c|c|c|c}
    \toprule
    \textbf{Dataset} & \textbf{Teacher} & \textbf{Student} & \textbf{Teacher-only / \ours} \\
    \midrule
    ImageNet-R & ViT-B/16 & ViT-B/16 & 49.11 / 52.27 \\
    ImageNet-R & ViT-B/16 & ViT-L/16 & 49.11 / 54.08 \\
    ImageNet-R & ViT-L/16 & ViT-B/16 & 54.42 / 58.18 \\
    ImageNet-R & ViT-L/16 & ViT-L/16 & 54.42 / 57.79 \\
    \midrule
    ImageNet-Sketch & ViT-B/16 & ViT-B/16 & 38.85 / 41.46 \\
    ImageNet-Sketch & ViT-B/16 & ViT-L/16 & 38.85 / 42.89 \\
    ImageNet-Sketch & ViT-L/16 & ViT-B/16 & 43.83 / 47.61 \\
    ImageNet-Sketch & ViT-L/16 & ViT-L/16 & 43.83 / 45.91 \\
    \bottomrule
    \end{tabular}}
\end{minipage}
\hfill
\begin{minipage}{0.42\textwidth}
    \centering
    \caption{Performance of \ours, BAN, ReviewKD, DKD on CIFAR100.}
    \label{Table_performance_comparison}
    \resizebox{0.98\linewidth}{!}{
    \begin{tabular}{c|c|c|c}
    \toprule
    \textbf{Method} & \textbf{Teacher} & \textbf{Student} & \textbf{Accuracy} \\
    \midrule
    Teacher-only & ResNet-50 & N/A & 84.09 \\
    BAN & ResNet-50 & ResNet-50 & 84.73 \\
    ReviewKD & ResNet-50 & ResNet-50 & 85.31 \\
    DKD & ResNet-50 & ResNet-50 & 85.17 \\
    \ours & ResNet-50 & ResNet-50 & 86.04 \\
    \midrule
    Teacher-only & ViT-B/16 & N/A & 91.57 \\
    BAN & ViT-B/16 & ViT-B/16 & 92.44 \\
    ReviewKD & ViT-B/16 & ViT-B/16 & 92.73 \\
    DKD & ViT-B/16 & ViT-B/16 & 92.82 \\
    \ours & ViT-B/16 & ViT-B/16 & 93.17 \\
    \bottomrule
    \end{tabular}}
\end{minipage}
}
\end{table}

\paragraph{Results on Validation Datasets.} We provide additional results on the official validation datasets for PTB and WikiText-103 in Table \ref{Table_perplexity_comparison}. These results demonstrate that \ours consistently outperforms the Teacher-only approach on both the validation and test datasets for PTB and WikiText-103, further validating the effectiveness of \ours.

\begin{table}[!htb]
\centering
\caption{Test/Validation perplexity of \ours and Teacher-only on the official test/validation datasets.}
\label{Table_perplexity_comparison}
\resizebox{1\textwidth}{!}{
\begin{tabular}{c|c|c|c|c|c|c}
\toprule
\textbf{Dataset} & \textbf{Teacher} & \textbf{Student} & \textbf{Teacher-only (Valid)} & \textbf{Teacher-only (Test)} & \textbf{\ours (Valid)} & \textbf{\ours (Test)} \\
\midrule
PTB & LSTM & LSTM & 86.02 & 82.75 & 73.98 & 71.72 \\
PTB & AWD-LSTM & AWD-LSTM & 60.62 & 58.69 & 55.07 & 53.31 \\
Wikitext-103 & Transformer-XL-B & Transformer-XL-B & 24.68 & 23.72 & 22.24 & 21.65 \\
Wikitext-103 & Transformer-XL-L & Transformer-XL-L & 18.65 & 18.50 & 16.41 & 16.47 \\
\bottomrule
\end{tabular}}
\end{table}

\paragraph{Performance of Student Models.} We present the results for the student models in Table \ref{Table_student_performance}. Our observations indicate that when the student and teacher models share the same architecture, the student models can achieve performance levels comparable to those of the teacher models. While the performance of the student models improves under \ours, it is important to highlight that \ours is primarily designed to enhance the generalization capabilities of the teacher model.

\begin{table}[!tb]
\centering
\caption{The performance of student models in LoT on language modeling and image classification.}
\label{Table_student_performance}
\resizebox{\textwidth}{!}{
\begin{tabular}{c|c|c|c|c|c|c}
\toprule
\textbf{Task} & \textbf{Dataset} & \textbf{Teacher} & \textbf{Student} & \textbf{Teacher-only} & \textbf{\ours (Teacher)} & \textbf{\ours (Student)} \\
\midrule
Language Modeling & PTB & LSTM & LSTM & 82.75 & 71.72 & 73.33 \\
Language Modeling & WikiText-103 & Transformer-XL-L & Transformer-XL-L & 18.50 & 16.47 & 16.89 \\
\midrule
Image Classification & CIFAR100 & ResNet-50 & ResNet-18 & 84.09 & 85.77 & 83.24 \\
Image Classification & CIFAR100 & ResNet-50 & ResNet-50 & 84.09 & 86.04 & 85.72 \\
Image Classification & ImageNet-1K & ViT-B/16 & ViT-B/16 & 91.57 & 93.17 & 92.95 \\
Image Classification & ImageNet-1K & ViT-B/16 & ViT-L/16 & 91.57 & 93.25 & 93.89 \\
\bottomrule
\end{tabular}}
\end{table}

\paragraph{Detialed Computation Cost.} We provide a detailed comparison of the computational budget for \ours and Teacher-only in Table \ref{Table_computational_resources}. Our analysis shows that \ours uses the same number of CPU cores as Teacher-only, with GPU usage being 12\% to 55\% higher. Despite this, \ours exhibits lower training times compared to Teacher-only (except in RL tasks) when subjected to the same total training epochs/steps, while still achieving significant performance improvements.

\begin{table}[!tb]
\centering
\caption{Computational resources, memory usage, and training time of \ours and Teacher-only.}
\label{Table_computational_resources}
\resizebox{\textwidth}{!}{
\begin{tabular}{c|c|c|c|c|c|c|c}
\toprule
\textbf{Dataset} & \textbf{Teacher Model / Student Model} & \textbf{Total Train Steps} & \textbf{Computational Resources} & \textbf{CPU Usage} & \textbf{GPU Usage} & \textbf{Training Time} & \textbf{Performance} \\
& & \textbf{(teacher+student)} & & \textbf{(Teacher-only/\ours)} & \textbf{(Teacher-only/LoT)} & \textbf{(Teacher-only/\ours)} & \textbf{(Teacher-only/\ours)} \\
\midrule
BeamRider & Standard Network / Standard Network & 20M frames & 1 NVIDIA A6000 48GB GPU & 16 core / 16 core & 0.8 GB / 0.9 GB & 10 h / 10.1 h & 3,651 score / 5,956 score ($\uparrow$) \\
PTB & LSTM / LSTM & 60 epochs & 1 $\times$ NVIDIA A100 40GB GPU & 1 core / 1 core & 1.1 GB / 1.5 GB & 0.6 h / 0.3 h & 82.8 ppl / 71.7 ppl ($\downarrow$) \\
WikiText-103 & Transformer-XL-L / Transformer-XL-L & 0.3M steps & 4 $\times$ NVIDIA A100 40GB GPU & 4 core / 4 core & 4 $\times$ 21.4 GB / 4 $\times$ 33.2 GB & 85.6 h / 67.7 h & 18.5 ppl / 16.5 ppl ($\downarrow$) \\
GSM8K & LLaMA-2 7B / LLaMA-2 7B & 4 epochs & 8 $\times$ NVIDIA A100 40GB GPU & 8 core / 8 core & 8 $\times$ 27.4 GB / 8 $\times$ 39.8 GB & 8.1 h / 6.7 h & 39.8 acc / 41.9 acc ($\uparrow$) \\
CIFAR100 & ResNet-50 / ResNet-18 & 60 epochs & 1 $\times$ NVIDIA A100 40GB GPU & 1 core / 1 core & 13.6 GB / 16.7 GB & 0.7 h / 0.5 h & 84.1 acc / 85.8 acc ($\uparrow$) \\
ImageNet-1K & ViT-L/16 / ViT-B/16 & 20K steps & 4 $\times$ NVIDIA A100 40GB GPU & 4 core / 4 core & 4 $\times$ 17.5 GB / 4 $\times$ 23.1 GB & 28.9 h / 18.7 h & 85.2 acc / 86.0 acc ($\uparrow$) \\
\bottomrule
\end{tabular}
}
\end{table}

\paragraph{Ablation of Metrics in \ours Regularizer.} We conduct experiments with different metrics for the “imitability” measurement, such as L2 loss. However, we find that using KL-divergence achieves better performance compared to L2 loss. The results of utilizing L2 loss for the \ours regularizer with ViT-B/16 and ViT-L/16 on CIFAR-100 are presented in Table \ref{Table_L2_loss_performance}. These results show that using L2 loss for the \ours regularizer also brings performance improvements, further indicating the effectiveness of \ours regularization.

\begin{table}[!htb]
\centering
\caption{Performance of using L2 loss for the \ours regularizer on CIFAR100.}
\label{Table_L2_loss_performance}
\resizebox{0.8\textwidth}{!}{
\begin{tabular}{c|c|c|c|c|c}
\toprule
\textbf{Dataset} & \textbf{Teacher} & \textbf{Student} & \textbf{Teacher-only} & \textbf{\ours (KL-Divergence)} & \textbf{\ours (L2)} \\
\midrule
CIFAR100 & ViT-B/16 & ViT-B/16 & 91.57 & 93.17 & 92.77 \\
CIFAR100 & ViT-B/16 & ViT-L/16 & 91.57 & 93.25 & 92.94 \\
CIFAR100 & ViT-L/16 & ViT-B/16 & 93.44 & 94.29 & 94.12 \\
CIFAR100 & ViT-L/16 & ViT-L/16 & 93.44 & 94.18 & 94.05 \\
\bottomrule
\end{tabular}}
\end{table}

\newpage
\section*{NeurIPS Paper Checklist}

\begin{enumerate}

\item {\bf Claims}
    \item[] Question: Do the main claims made in the abstract and introduction accurately reflect the paper's contributions and scope?
    \item[] Answer: \answerYes{} 
    \item[] Justification: The Abstract and Introduction (Section~\ref{section_introduction}) in this paper reflect the contributions of our method. The strong results in our experiments further reflect our contributions.
    \item[] Guidelines:
    \begin{itemize}
        \item The answer NA means that the abstract and introduction do not include the claims made in the paper.
        \item The abstract and/or introduction should clearly state the claims made, including the contributions made in the paper and important assumptions and limitations. A No or NA answer to this question will not be perceived well by the reviewers. 
        \item The claims made should match theoretical and experimental results, and reflect how much the results can be expected to generalize to other settings. 
        \item It is fine to include aspirational goals as motivation as long as it is clear that these goals are not attained by the paper. 
    \end{itemize}

\item {\bf Limitations}
    \item[] Question: Does the paper discuss the limitations of the work performed by the authors?
    \item[] Answer: \answerYes{} 
    \item[] Justification: The potential limitation of \ours lies in the additional computational and memory costs required for training the student models. However, we demonstrate that \ours achieves better generalization with fewer training steps compared to Teacher-only models and the flexibility in choosing student models can accommodate varying resource constraints. We provide discussions in Section~\ref{section_cost_analysis} and Appendix~\ref{appendix_limitation}.
    \item[] Guidelines:
    \begin{itemize}
        \item The answer NA means that the paper has no limitation while the answer No means that the paper has limitations, but those are not discussed in the paper. 
        \item The authors are encouraged to create a separate "Limitations" section in their paper.
        \item The paper should point out any strong assumptions and how robust the results are to violations of these assumptions (e.g., independence assumptions, noiseless settings, model well-specification, asymptotic approximations only holding locally). The authors should reflect on how these assumptions might be violated in practice and what the implications would be.
        \item The authors should reflect on the scope of the claims made, e.g., if the approach was only tested on a few datasets or with a few runs. In general, empirical results often depend on implicit assumptions, which should be articulated.
        \item The authors should reflect on the factors that influence the performance of the approach. For example, a facial recognition algorithm may perform poorly when image resolution is low or images are taken in low lighting. Or a speech-to-text system might not be used reliably to provide closed captions for online lectures because it fails to handle technical jargon.
        \item The authors should discuss the computational efficiency of the proposed algorithms and how they scale with dataset size.
        \item If applicable, the authors should discuss possible limitations of their approach to address problems of privacy and fairness.
        \item While the authors might fear that complete honesty about limitations might be used by reviewers as grounds for rejection, a worse outcome might be that reviewers discover limitations that aren't acknowledged in the paper. The authors should use their best judgment and recognize that individual actions in favor of transparency play an important role in developing norms that preserve the integrity of the community. Reviewers will be specifically instructed to not penalize honesty concerning limitations.
    \end{itemize}

\item {\bf Theory Assumptions and Proofs}
    \item[] Question: For each theoretical result, does the paper provide the full set of assumptions and a complete (and correct) proof?
    \item[] Answer: \answerNA{} 
    \item[] Justification: The paper does not include theoretical results
    \item[] Guidelines:
    \begin{itemize}
        \item The answer NA means that the paper does not include theoretical results. 
        \item All the theorems, formulas, and proofs in the paper should be numbered and cross-referenced.
        \item All assumptions should be clearly stated or referenced in the statement of any theorems.
        \item The proofs can either appear in the main paper or the supplemental material, but if they appear in the supplemental material, the authors are encouraged to provide a short proof sketch to provide intuition. 
        \item Inversely, any informal proof provided in the core of the paper should be complemented by formal proofs provided in appendix or supplemental material.
        \item Theorems and Lemmas that the proof relies upon should be properly referenced. 
    \end{itemize}

    \item {\bf Experimental Result Reproducibility}
    \item[] Question: Does the paper fully disclose all the information needed to reproduce the main experimental results of the paper to the extent that it affects the main claims and/or conclusions of the paper (regardless of whether the code and data are provided or not)?
    \item[] Answer: \answerYes{} 
    \item[] Justification: The implementation details of this paper is fully illustrated in Appendix~\ref{Appendix_implementation_details} and we make an extensive effort to ensure the reproducibility of the results in this paper.
    \item[] Guidelines:
    \begin{itemize}
        \item The answer NA means that the paper does not include experiments.
        \item If the paper includes experiments, a No answer to this question will not be perceived well by the reviewers: Making the paper reproducible is important, regardless of whether the code and data are provided or not.
        \item If the contribution is a dataset and/or model, the authors should describe the steps taken to make their results reproducible or verifiable. 
        \item Depending on the contribution, reproducibility can be accomplished in various ways. For example, if the contribution is a novel architecture, describing the architecture fully might suffice, or if the contribution is a specific model and empirical evaluation, it may be necessary to either make it possible for others to replicate the model with the same dataset, or provide access to the model. In general. releasing code and data is often one good way to accomplish this, but reproducibility can also be provided via detailed instructions for how to replicate the results, access to a hosted model (e.g., in the case of a large language model), releasing of a model checkpoint, or other means that are appropriate to the research performed.
        \item While NeurIPS does not require releasing code, the conference does require all submissions to provide some reasonable avenue for reproducibility, which may depend on the nature of the contribution. For example
        \begin{enumerate}
            \item If the contribution is primarily a new algorithm, the paper should make it clear how to reproduce that algorithm.
            \item If the contribution is primarily a new model architecture, the paper should describe the architecture clearly and fully.
            \item If the contribution is a new model (e.g., a large language model), then there should either be a way to access this model for reproducing the results or a way to reproduce the model (e.g., with an open-source dataset or instructions for how to construct the dataset).
            \item We recognize that reproducibility may be tricky in some cases, in which case authors are welcome to describe the particular way they provide for reproducibility. In the case of closed-source models, it may be that access to the model is limited in some way (e.g., to registered users), but it should be possible for other researchers to have some path to reproducing or verifying the results.
        \end{enumerate}
    \end{itemize}

\item {\bf Open access to data and code}
    \item[] Question: Does the paper provide open access to the data and code, with sufficient instructions to faithfully reproduce the main experimental results, as described in supplemental material?
    \item[] Answer: \answerYes{} 
    \item[] Justification: We provide the code of this paper in the additional supplementary material.
    \item[] Guidelines:
    \begin{itemize}
        \item The answer NA means that paper does not include experiments requiring code.
        \item Please see the NeurIPS code and data submission guidelines (\url{https://nips.cc/public/guides/CodeSubmissionPolicy}) for more details.
        \item While we encourage the release of code and data, we understand that this might not be possible, so “No” is an acceptable answer. Papers cannot be rejected simply for not including code, unless this is central to the contribution (e.g., for a new open-source benchmark).
        \item The instructions should contain the exact command and environment needed to run to reproduce the results. See the NeurIPS code and data submission guidelines (\url{https://nips.cc/public/guides/CodeSubmissionPolicy}) for more details.
        \item The authors should provide instructions on data access and preparation, including how to access the raw data, preprocessed data, intermediate data, and generated data, etc.
        \item The authors should provide scripts to reproduce all experimental results for the new proposed method and baselines. If only a subset of experiments are reproducible, they should state which ones are omitted from the script and why.
        \item At submission time, to preserve anonymity, the authors should release anonymized versions (if applicable).
        \item Providing as much information as possible in supplemental material (appended to the paper) is recommended, but including URLs to data and code is permitted.
    \end{itemize}

\item {\bf Experimental Setting/Details}
    \item[] Question: Does the paper specify all the training and test details (e.g., data splits, hyperparameters, how they were chosen, type of optimizer, etc.) necessary to understand the results?
    \item[] Answer: \answerYes{} 
    \item[] Justification: The experimental setting and details are introduced in Appendix~\ref{Appendix_implementation_details}.
    \item[] Guidelines:
    \begin{itemize}
        \item The answer NA means that the paper does not include experiments.
        \item The experimental setting should be presented in the core of the paper to a level of detail that is necessary to appreciate the results and make sense of them.
        \item The full details can be provided either with the code, in appendix, or as supplemental material.
    \end{itemize}

\item {\bf Experiment Statistical Significance}
    \item[] Question: Does the paper report error bars suitably and correctly defined or other appropriate information about the statistical significance of the experiments?
    \item[] Answer: \answerYes{} 
    \item[] Justification: All our main results are averaged over multiple runs and the error bar are provided in our results.
    \item[] Guidelines:
    \begin{itemize}
        \item The answer NA means that the paper does not include experiments.
        \item The authors should answer "Yes" if the results are accompanied by error bars, confidence intervals, or statistical significance tests, at least for the experiments that support the main claims of the paper.
        \item The factors of variability that the error bars are capturing should be clearly stated (for example, train/test split, initialization, random drawing of some parameter, or overall run with given experimental conditions).
        \item The method for calculating the error bars should be explained (closed form formula, call to a library function, bootstrap, etc.)
        \item The assumptions made should be given (e.g., Normally distributed errors).
        \item It should be clear whether the error bar is the standard deviation or the standard error of the mean.
        \item It is OK to report 1-sigma error bars, but one should state it. The authors should preferably report a 2-sigma error bar than state that they have a 96\% CI, if the hypothesis of Normality of errors is not verified.
        \item For asymmetric distributions, the authors should be careful not to show in tables or figures symmetric error bars that would yield results that are out of range (e.g. negative error rates).
        \item If error bars are reported in tables or plots, The authors should explain in the text how they were calculated and reference the corresponding figures or tables in the text.
    \end{itemize}

\item {\bf Experiments Compute Resources}
    \item[] Question: For each experiment, does the paper provide sufficient information on the computer resources (type of compute workers, memory, time of execution) needed to reproduce the experiments?
    \item[] Answer: \answerYes{} 
    \item[] Justification: The compute resources details are provided in Appendix~\ref{Appendix_implementation_details}.
    \item[] Guidelines:
    \begin{itemize}
        \item The answer NA means that the paper does not include experiments.
        \item The paper should indicate the type of compute workers CPU or GPU, internal cluster, or cloud provider, including relevant memory and storage.
        \item The paper should provide the amount of compute required for each of the individual experimental runs as well as estimate the total compute. 
        \item The paper should disclose whether the full research project required more compute than the experiments reported in the paper (e.g., preliminary or failed experiments that didn't make it into the paper). 
    \end{itemize}
    
\item {\bf Code Of Ethics}
    \item[] Question: Does the research conducted in the paper conform, in every respect, with the NeurIPS Code of Ethics \url{https://neurips.cc/public/EthicsGuidelines}?
    \item[] Answer: \answerYes{} 
    \item[] Justification: The paper conforms to the NeurIPS Code of Ethics.
    \item[] Guidelines:
    \begin{itemize}
        \item The answer NA means that the authors have not reviewed the NeurIPS Code of Ethics.
        \item If the authors answer No, they should explain the special circumstances that require a deviation from the Code of Ethics.
        \item The authors should make sure to preserve anonymity (e.g., if there is a special consideration due to laws or regulations in their jurisdiction).
    \end{itemize}

\item {\bf Broader Impacts}
    \item[] Question: Does the paper discuss both potential positive societal impacts and negative societal impacts of the work performed?
    \item[] Answer: \answerYes{} 
    \item[] Justification: The societal impacts of this paper is discussed in Appendix~\ref{appendix_social_impacts}.
    \item[] Guidelines:
    \begin{itemize}
        \item The answer NA means that there is no societal impact of the work performed.
        \item If the authors answer NA or No, they should explain why their work has no societal impact or why the paper does not address societal impact.
        \item Examples of negative societal impacts include potential malicious or unintended uses (e.g., disinformation, generating fake profiles, surveillance), fairness considerations (e.g., deployment of technologies that could make decisions that unfairly impact specific groups), privacy considerations, and security considerations.
        \item The conference expects that many papers will be foundational research and not tied to particular applications, let alone deployments. However, if there is a direct path to any negative applications, the authors should point it out. For example, it is legitimate to point out that an improvement in the quality of generative models could be used to generate deepfakes for disinformation. On the other hand, it is not needed to point out that a generic algorithm for optimizing neural networks could enable people to train models that generate Deepfakes faster.
        \item The authors should consider possible harms that could arise when the technology is being used as intended and functioning correctly, harms that could arise when the technology is being used as intended but gives incorrect results, and harms following from (intentional or unintentional) misuse of the technology.
        \item If there are negative societal impacts, the authors could also discuss possible mitigation strategies (e.g., gated release of models, providing defenses in addition to attacks, mechanisms for monitoring misuse, mechanisms to monitor how a system learns from feedback over time, improving the efficiency and accessibility of ML).
    \end{itemize}
    
\item {\bf Safeguards}
    \item[] Question: Does the paper describe safeguards that have been put in place for responsible release of data or models that have a high risk for misuse (e.g., pretrained language models, image generators, or scraped datasets)?
    \item[] Answer: \answerNA{} 
    \item[] Justification: This paper does not release new models or datasets.
    \item[] Guidelines:
    \begin{itemize}
        \item The answer NA means that the paper poses no such risks.
        \item Released models that have a high risk for misuse or dual-use should be released with necessary safeguards to allow for controlled use of the model, for example by requiring that users adhere to usage guidelines or restrictions to access the model or implementing safety filters. 
        \item Datasets that have been scraped from the Internet could pose safety risks. The authors should describe how they avoided releasing unsafe images.
        \item We recognize that providing effective safeguards is challenging, and many papers do not require this, but we encourage authors to take this into account and make a best faith effort.
    \end{itemize}

\item {\bf Licenses for existing assets}
    \item[] Question: Are the creators or original owners of assets (e.g., code, data, models), used in the paper, properly credited and are the license and terms of use explicitly mentioned and properly respected?
    \item[] Answer: \answerYes{}{} 
    \item[] Justification: We use publicly popular datasets and models and obtain the license of using LLaMA models. We credit the license and term of use in the code in the supplementary material.
    \item[] Guidelines:
    \begin{itemize}
        \item The answer NA means that the paper does not use existing assets.
        \item The authors should cite the original paper that produced the code package or dataset.
        \item The authors should state which version of the asset is used and, if possible, include a URL.
        \item The name of the license (e.g., CC-BY 4.0) should be included for each asset.
        \item For scraped data from a particular source (e.g., website), the copyright and terms of service of that source should be provided.
        \item If assets are released, the license, copyright information, and terms of use in the package should be provided. For popular datasets, \url{paperswithcode.com/datasets} has curated licenses for some datasets. Their licensing guide can help determine the license of a dataset.
        \item For existing datasets that are re-packaged, both the original license and the license of the derived asset (if it has changed) should be provided.
        \item If this information is not available online, the authors are encouraged to reach out to the asset's creators.
    \end{itemize}

\item {\bf New Assets}
    \item[] Question: Are new assets introduced in the paper well documented and is the documentation provided alongside the assets?
    \item[] Answer: \answerYes{}{} 
    \item[] Justification: We include license in the code in the supplementary material.
    \item[] Guidelines:
    \begin{itemize}
        \item The answer NA means that the paper does not release new assets.
        \item Researchers should communicate the details of the dataset/code/model as part of their submissions via structured templates. This includes details about training, license, limitations, etc. 
        \item The paper should discuss whether and how consent was obtained from people whose asset is used.
        \item At submission time, remember to anonymize your assets (if applicable). You can either create an anonymized URL or include an anonymized zip file.
    \end{itemize}

\item {\bf Crowdsourcing and Research with Human Subjects}
    \item[] Question: For crowdsourcing experiments and research with human subjects, does the paper include the full text of instructions given to participants and screenshots, if applicable, as well as details about compensation (if any)? 
    \item[] Answer: \answerNA{} 
    \item[] Justification: The paper does not involve crowdsourcing nor research with human subjects.
    \item[] Guidelines:
    \begin{itemize}
        \item The answer NA means that the paper does not involve crowdsourcing nor research with human subjects.
        \item Including this information in the supplemental material is fine, but if the main contribution of the paper involves human subjects, then as much detail as possible should be included in the main paper. 
        \item According to the NeurIPS Code of Ethics, workers involved in data collection, curation, or other labor should be paid at least the minimum wage in the country of the data collector. 
    \end{itemize}

\item {\bf Institutional Review Board (IRB) Approvals or Equivalent for Research with Human Subjects}
    \item[] Question: Does the paper describe potential risks incurred by study participants, whether such risks were disclosed to the subjects, and whether Institutional Review Board (IRB) approvals (or an equivalent approval/review based on the requirements of your country or institution) were obtained?
    \item[] Answer: \answerNA{} 
    \item[] Justification: The paper does not involve crowdsourcing nor research with human subjects.
    \item[] Guidelines:
    \begin{itemize}
        \item The answer NA means that the paper does not involve crowdsourcing nor research with human subjects.
        \item Depending on the country in which research is conducted, IRB approval (or equivalent) may be required for any human subjects research. If you obtained IRB approval, you should clearly state this in the paper. 
        \item We recognize that the procedures for this may vary significantly between institutions and locations, and we expect authors to adhere to the NeurIPS Code of Ethics and the guidelines for their institution. 
        \item For initial submissions, do not include any information that would break anonymity (if applicable), such as the institution conducting the review.
    \end{itemize}

\end{enumerate}

\end{document}